\documentclass{article}
\pdfoutput=1

\usepackage[final]{neurips_2018}

\usepackage[utf8]{inputenc} 
\usepackage[T1]{fontenc}    
\usepackage{hyperref}       
\usepackage{url}            
\usepackage{booktabs}       
\usepackage{amsfonts}       
\usepackage{amsmath}
\usepackage{nicefrac}       
\usepackage{microtype}      

\usepackage{xspace}
\usepackage{multirow}
\usepackage{boldline}
\usepackage{subfig}
\usepackage{array}
\usepackage{graphicx}
\usepackage{algorithm}
\usepackage{algorithmic}
\usepackage{color}
\usepackage{wrapfig}

\usepackage[numbers]{natbib}

\newcommand{\Skip}[1]{{}}

\newcommand{\ie}{\textit{i}.\textit{e}.\ }
\newcommand{\eg}{\textit{e}.\textit{g}.\ }
\newcommand{\myfig}[1]{Figure \ref{#1}}
\newcommand{\mytable}[1]{Table \ref{#1}}

\newcommand{\mysec}[1]{Section \ref{#1}}
\newcommand{\myalg}[1]{Algorithm \ref{#1}}
\newcommand{\vspaceAfterSection}{\vspace{-0.5em}}

\newcommand{\OurMethod}{\textsc{MuMoMAML}\xspace}
\newcommand{\SeqToSeq}{\textsc{Seq2Seq}\xspace}

\title{
Toward Multimodal Model-Agnostic Meta-Learning
}

\author{
Risto Vuorio$^1$ \hskip2em Shao-Hua Sun$^2$ \hskip2em Hexiang Hu$^2$ \hskip2em Joseph J. Lim$^2$ \\
SK T-Brain$^1$ \hskip2em University of Southern California$^2$ 
}

\begin{document}

\maketitle

\begin{abstract}
  Gradient-based meta-learners such as MAML~\cite{finn_model-agnostic_2017} are able to learn a meta-prior from similar tasks to adapt to novel tasks from the same distribution with few gradient updates.
  One important limitation of such frameworks is that they seek a common initialization shared across the entire task distribution, substantially limiting the diversity of the task distributions that they are able to learn from.
  In this paper, we augment MAML
  with the capability to identify tasks sampled from a multimodal task distribution and adapt quickly through gradient updates.
  Specifically, we propose a multimodal MAML algorithm that is able to modulate its meta-learned prior according to the identified task, allowing faster adaptation.
  We evaluate the proposed model on a diverse set of problems including regression, few-shot image classification, and reinforcement learning.
  The results demonstrate the effectiveness of our model in modulating the meta-learned prior in response to the characteristics of tasks sampled from a multimodal distribution.
\end{abstract}

\section{Introduction}
\label{sec:intro}
\vspaceAfterSection
Recent advances in meta-learning offer machines a way to learn from
a distribution of tasks and adapt to a new task from the same
distribution using few samples~\cite{koch2015siamese, vinyals2016matching}.
Different approaches for engaging the task distribution exist.
Optimization-based meta-learning methods offer learnable learning rules and
optimization algorithms~\cite{schmidhuber:1987:srl, bengio1992optimization, ravi_optimization_2016, andrychowicz_learning_2016, hochreiter2001learning},
metric-based meta learners~\cite{koch2015siamese, vinyals2016matching, snell_prototypical_2017, shyam2017attentive, Sung_2018_CVPR}
address few-shot classification by encoding task-related knowledge in a learned metric space. 
Model-based meta-learning approaches~\citep{duan2016rl, wang2016learning, munkhdalai_meta_2017, mishra_simple_2017} generalize to a wider range of learning scenarios,
seeking to recognize the task identity from a few data samples and adapt to the tasks
by adjusting a model's state (\eg RNN's internal states).
Model-based methods demonstrate high performance at the expense of hand-designing architectures, yet the optimal strategy of designing a meta-learner for arbitrary tasks may not be obvious to humans.
On the other hand, model-agnostic gradient-based meta-learners~\citep{finn_model-agnostic_2017, finn_probabilistic_2018, kim_bayesian_2018, lee_gradient-based_2018, grant_recasting_2018}
seek an initialization of model parameters 
such that a small number of gradient updates will lead to fast learning on a new task, offering the flexibility in the choice of models. 

While most existing gradient-based meta-learners rely on a single initialization,
different modes of a task distribution can require substantially different parameters,
making it infeasible to find a common initialization point for all tasks, given the same adaptation routine.
When the modes of a task distribution are disjoint and far apart, one can imagine that a set of separate
meta-learners with each covering one mode could better master the full distribution.
However, this not only 
requires additional identity information about the modes, 
which is not always available or could be ambiguous when the task modes are not clearly disjoint,
but also eliminates the possibility of associating transferable knowledge across different modes of a task distribution.
To overcome this issue,
we aim to develop a meta-learner that  
acquires a prior over 
a multimodal task distribution and adapts quickly 
within the distribution with gradient descent.

To this end, we leverage the strengths of the two main lines of existing meta-learning methods: model-based and gradient-based meta-learning.
Specifically, we propose to augment gradient based meta-learners with 
the capability of generalizing across a multimodal task distribution. 
Instead of learning a single initialization point in the parameter space, 
we propose to first estimate the mode of a sampled task by examining task related samples.
Given the estimated task mode, our model then performs a step of \textit{model-based adaptation}
to modulate the meta-learned prior in the parameter space to fit the sampled task.
Then, from this model adapted meta-prior, a few steps of
\textit{gradient-based adaptation} are performed towards the target task
to progressively improve the performance on the task.
This main idea is illustrated in \myfig{fig:model}. 

\Skip{
Optimization-based meta-learning approaches learn to optimize a learner model by optimizing learned parameters~\cite{schmidhuber:1987:srl, bengio1992optimization}, 
representing a learning as a reinforcement learning policy~\cite{li_learning_2016},
or learning update rules~\cite{andrychowicz_learning_2016, ravi_optimization_2016}.
Metric-based meta learners strive to learn a metric which can be used to compare two different samples effectively and perform tasks in the learned metric space~\cite{koch2015siamese, vinyals2016matching, snell_prototypical_2017, shyam2017attentive, Sung_2018_CVPR}.
Model-based meta-learning frameworks learn to recognize the identities of tasks and adjust the model state (\eg the internal state of an RNN) to fit the task~\cite{santoro_meta-learning_2016, duan2016rl, munkhdalai_meta_2017}.
While achieving impressive results on a wide range of tasks, those methods are not able to continually improve the performance during the meta-testing phase.
\cite{finn_model-agnostic_2017} and its extensions ~\cite{finn_probabilistic_2018, kim_bayesian_2018, lee_gradient-based_2018, grant_recasting_2018}, known as gradient-based meta-learning, aim to estimate a parameter initialization that provides a favorable inductive bias for fast adaptation.
However, assuming tasks are sampled from a concentrated distribution and pursuing a common initialization to all tasks can substantially limit the performance of such methods on multimodal task distributions where the center in the task space becomes ambiguous.
}

\Skip{
\begin{figure*}[t]
    \centering
    \begin{tabular}{cc}
    \includegraphics[height=130pt]{figures/mumomaml_teaser_2.pdf} &
    \includegraphics[height=130pt]{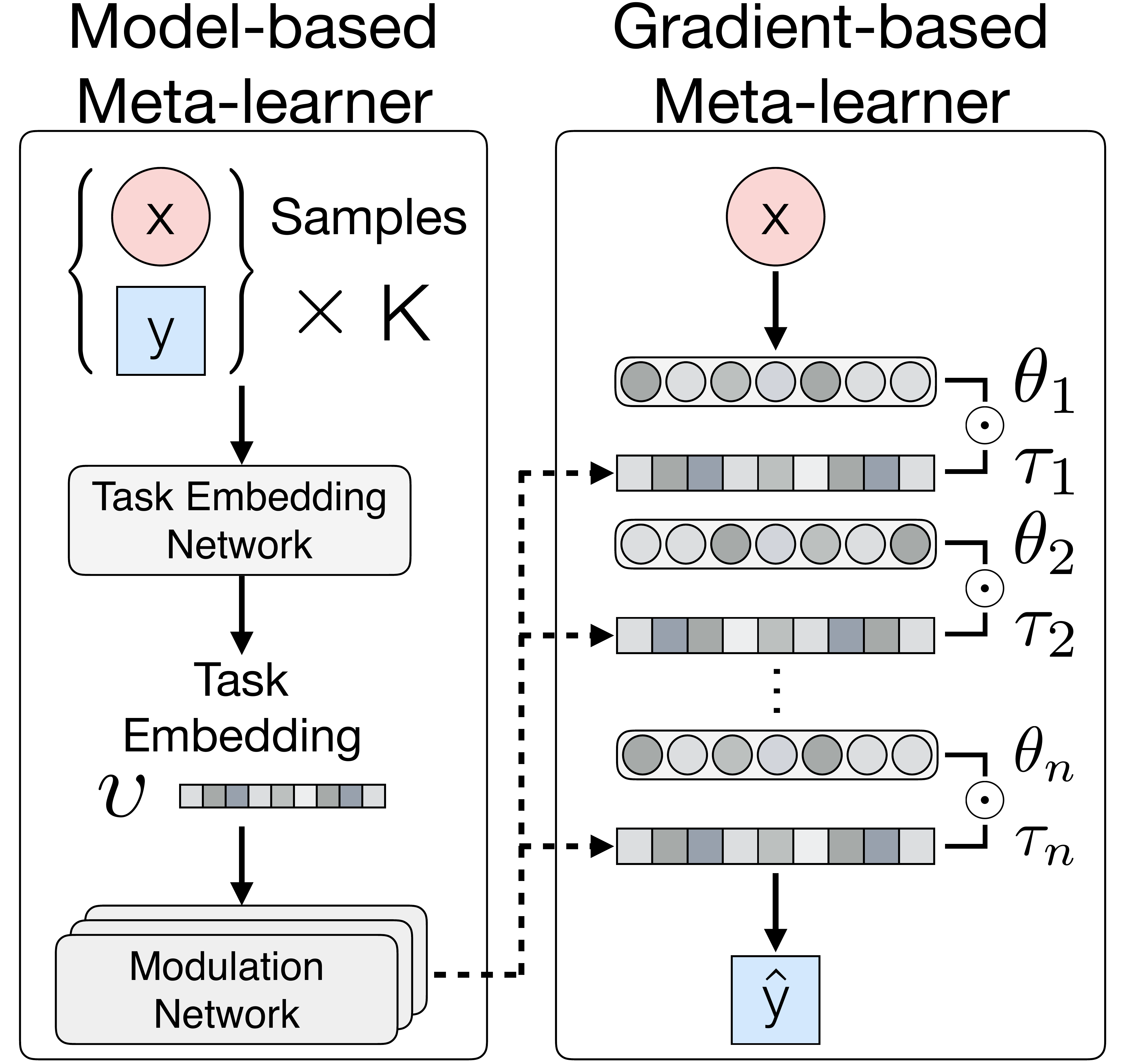} \\
    (a) & (b)\\
    \end{tabular}
    
    \caption{
    (a) Diagram of multimodal model-agnostic meta-learning (\OurMethod) algorithm, which aims to acquire a meta-learned prior parameterized by $\theta$ and $\omega$, enabling fast adaptation to a task sampled from a multimodal distribution.
    Areas of different color indicate modes of tasks.
    (b) Overview of our model. Our model consists of two components: a model-based and gradient-based meta-learner. The former strives to identify the mode of a task distribution from a few samples and modulate the prior accordingly; the latter performs gradient updates to effectively yield better performance.
        \label{fig:illustration}
    }
\end{figure*}
}

\vspaceAfterSection
\section{Method}
\label{sec:method}
\begin{figure*}[ttt!]

\begin{tabular}{cc}
\begin{minipage}{.36\textwidth}
    
    \centering
    \includegraphics[width=\textwidth]{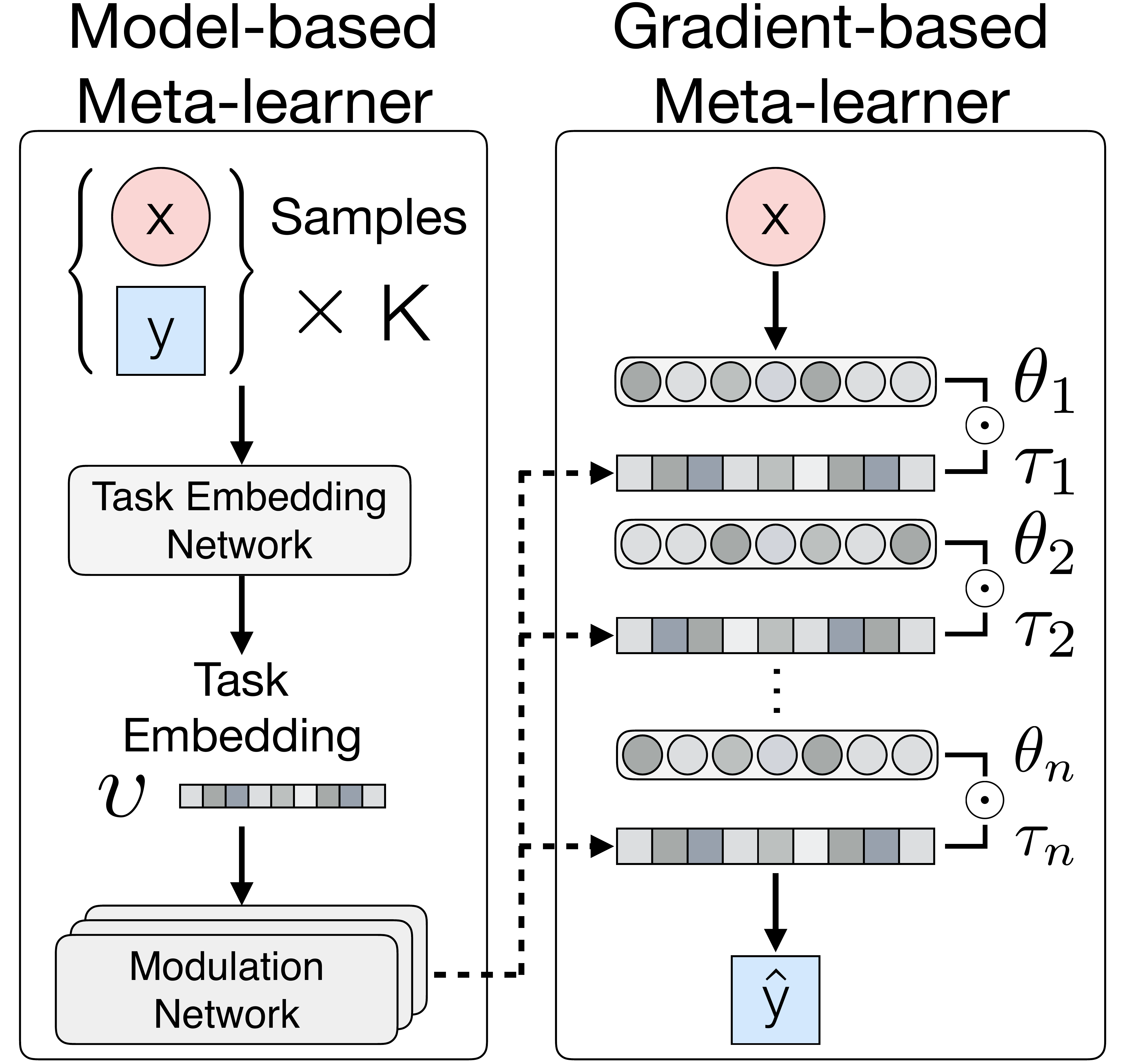}
    \captionof{figure}{Model overview.} \label{fig:model}


\end{minipage} &

\hfill

\scalebox{0.81}{
    \begin{minipage}{.72\textwidth}
    \vspace{-20pt}
    \begin{algorithm}[H]
    \captionof{algorithm}{\textsc{Meta-Training Procedure.}} \label{alg:train}
        \begin{algorithmic}[1]
        \STATE {\bfseries Input:} Task distribution $P(\mathcal{T})$, Hyper-parameters $\alpha$ and $\beta$
    		\STATE Randomly initialize $\theta$ and $\omega$.
    		\WHILE{not DONE}
        		\STATE Sample batches of tasks $\mathcal{T}_j \sim P(\mathcal{T})$
        		\FOR{$\mathbf{all}$ j}
        		    \STATE Infer $\tau = g(\{x, y\}; \omega)$ with K samples from $\mathcal{D}^{\text{train}}_{\mathcal{T}_j}$
        		    \STATE Evaluate $\nabla_{\theta} \mathcal{L}_{\mathcal{T}_j}\Big(f(x ; \theta, \tau);\mathcal{D}^{\text{train}}_{\mathcal{T}_j}\Big)$ 
        		    w.r.t the K samples
        		    \STATE Compute adapted parameter with gradient descent: \\ $\theta_{\mathcal{T}_j}' = \theta - \alpha \nabla_{\theta} \mathcal{L}_{\mathcal{T}_j}\big(f(x ; \theta, \tau); \mathcal{D}^{\text{train}}_{\mathcal{T}_j}\big)$
        		\ENDFOR
        		\STATE Update $\theta \leftarrow \theta - \beta \nabla_{\theta} \sum_{T_j \sim P(\mathcal{T})} \mathcal{L}_{\mathcal{T}_j}\big(f(x ; \theta', \tau); \mathcal{D}^{\text{val}}_{\mathcal{T}_j}\big)$
        		\STATE Update $\omega \leftarrow \omega - \beta \nabla_{\omega} \sum_{T_j \sim P(\mathcal{T})} \mathcal{L}_{\mathcal{T}_j}\big(f(x ; \theta', \tau); \mathcal{D}^{\text{val}}_{\mathcal{T}_j}\big)$
    		\ENDWHILE  
    \end{algorithmic}
    
    \end{algorithm}
    
    \end{minipage}
}

\end{tabular}

\end{figure*}

\vspaceAfterSection
We aim to develop a Multi-Modal Model-Agnostic Meta-Learner (\OurMethod) that is able to 
quickly master a novel task sampled from \textit{a multimodal task distribution}.
To this end, we propose to leverage 
the ability of model-based meta-learners to identify the modes of a task distribution as well as 
the ability of gradient-based meta-learners to consistently improve the performance with a few gradient steps. 
Specifically, we propose to learn a model-based meta-learner 
that produces a set of task specific parameters to modulate the meta-learned prior parameters. 
Then, this modulated prior learns to adapt to a target task rapidly through gradient-based optimization. 
An illustration of our model is shown in \myfig{fig:model}.

The \textbf{gradient-based meta-learner}, parameterized by $\theta$, is optimized to quickly adapt to target tasks with few gradient steps by seeking a good parameter
initialization similar to~\cite{finn_model-agnostic_2017}.
For the architecture of the gradient-based meta-learner, we consider a neural network consisting of $N$
blocks where the $i$-th block is a convolutional or a fully-connected layer
parameterized by $\theta_i$.
The \textbf{model-based meta-learner}, parameterized by $\omega$, aims to identify the
mode of a sampled task from a few samples 
and then modulate the meta-learned prior parameters of the gradient-based
meta-learner to enable rapid adaptation in the identified mode. The model-based meta-learner consists of a \textit{task embedding network} and a \textit{modulation network}.

Given $K$ data points and labels $\{x_k, y_k\}_{k=1,...,K}$, 
the task embedding network $f$ learns to produce an embedding vector $\upsilon$
that encodes the characteristics of a task according to
$\upsilon = f(\{x_k, y_k\}_{k=1,...,K}; \omega_f)$. 
The modulation network $g$ learns to modulate the meta-learned prior of the
gradient-based meta-learner in the parameter space based on the task
embedding vector $\upsilon$.
To enable specialization of each block of the gradient-based meta-learner to the task,
we apply the modulation block-wise to activate or deactivate the units of a block
(\ie a channel of a convolutional layer or a neuron of a fully-connected layer).
Specifically, modulation network produces the modulation vectors for each block $i$
by $\tau_1, ..., \tau_N = g(\upsilon; \omega_{g})$, 
forming a collection of modulated parameters $\tau$.
We formalize the procedure of applying modulation as: $\phi_i = \theta_i \odot \tau_i$,
where $\phi_i$ represents the modulated prior parameters for the gradient-based meta-learner, 
and $\odot$ represents a general modulation function. 
In the experiments, we investigate some representative modulation operations
including attention-based modulation~\citep{mnih2014recurrent,vaswani_attention_2017} 
and feature-wise linear modulation (FiLM)~\citep{perez_film:_2017}. 

\noindent \textbf{Training}
\Skip{
During the meta-training phase, 
the model-based and gradient-based meta-learners
are trained jointly.
Given a batch of sampled tasks $\mathcal{T}_i$
the task embedding network produces embedding vectors $\upsilon$ that capture the semantics of the tasks.
With the embeddings, the modulation network produces a set of vectors $\tau_1, ..., \tau_N$, which modulate (\ie activate or deactivate) the gradient-based meta-learner in the parameter space.
Then, gradient based adaptation is performed to optimize our model to yield improvement on the tasks.
}
The training procedure for jointly optimizing the model-based and 
gradient-based meta-learners is summarized in \myalg{alg:train}.
Note that $\tau$ is not updated in the inner loop, 
as the model-based meta-learner is only responsible for finding a good task-specific initialization through modulation.
The implementation details can be found in \mysec{subsec:implementation} and \mysec{sec:experimental_details}. 

\Skip{
During the meta-testing phase, with a sampled task $\mathcal{T}_i$ and a few data samples $\mathcal{D}^{\text{train}}_{\mathcal{T}_i}$, we perform an adaptation step that corresponds to the inner loop of the meta-training. 
Our model-based meta-learner first identifies the mode a target task belongs to 
based on the input data samples and then infers the parameter $\tau$ to modulate the prior model $\theta$. Next, we perform gradient updates on the prior model $f(x; \theta, \tau)$ and compute the adapted parameters $\theta'$. Finally, we evaluate our adapted model $f(x; \theta', \tau)$ on the meta-test set to measure the performance of the model.
}

\Skip{
Supposing we have K data points $\{x_k, y_k\}_{k=1,...,K}$ sampled from $D^{train}$, the task embedding model encodes the training data points into a vector that encodes the characteristics of a task. Specifically, we learn a generator model $g$ to take a sequence of paired data $x$ and its labels $y$ as input, and generate the task-specific parameters as: $\tau_i = g_i(\{x, y\}_{k=1,...,K}; \omega)$. 
Note that our modulation is applied block-wise to the gradient-based meta-learner. We use an index $i$ to represent the task-specific parameters $\tau$ for a block $i$. A block in the neural network might contain one or more layers, which is parameterized by $\theta_i$. In the rest part of this paper, we omit the index $i$ as the default configuration for brevity. 

During this process, the task encoder model takes few shots of data and embeds them into an intermediate representation $\upsilon$ that summarizes the task characteristics. Next, multiple generation functions take this shared task representation $\upsilon$ as input and predict the task-specific parameters $\tau$ for each block in the gradient-based meta-learner model. 
Based on these block-wise parameters $\tau$, a variety of modulation operations can be used for updating the task-specific prior information in parameter $\theta$, in a block-wise manner. We formalize this procedure as: $\phi = \theta \odot \tau$,
where $\phi$ represents the modulated prior parameters for gradient-based meta-learner, and $\odot$ represents a general modulation function. 
In this paper, we investigated some of the most representative modulation operations, including attention-based modulation~\citep{mnih2014recurrent,vaswani_attention_2017} and feature-wise linear modulation (FiLM)~\citep{perez_film:_2017}. 
}

\vspaceAfterSection
\section{Experiments}
\label{sec:experiments}
\vspaceAfterSection
To verify that the proposed method is able to quickly master tasks sampled from multimodal task distributions,
we compare it with baselines on a variety of tasks, including regression, reinforcement learning, and few-shot image classification
\footnote{Due to the page limit, the results of few-shot image classification are presented and discussed in \mysec{sec:few-shot}}.

\begin{table}[t]
    \centering
    \small
    \caption{\small Model's performance on the \textbf{multimodal 5-shot regression} with two or three modes. Gaussian noise with $\mu=0$ and $\sigma=0.3$ is applied. The three mode regression is in general more difficult (thus higher error). In Multi-MAML, the GT modulation represents using ground-truth task identification to select different MAML models for each task mode. \OurMethod (wt. FiLM) outperforms other methods by a significant margin.}
    \scalebox{0.85}{\begin{tabular}{cc|cc|cc} 
    \multicolumn{2}{c|}{\textbf{Configuration}} & 
    \multicolumn{2}{c|}{\textbf{Two Modes (MSE)}} &
    \multicolumn{2}{c}{\textbf{Three Modes (MSE)}}\\ 
    Method & Modulation 
    & Post Modulation & Post Adaptation 
    & Post Modulation & Post Adaptation \\ \hlineB{4}
    MAML~\citep{finn_model-agnostic_2017} & - & 15.9255 & 1.0852 & 12.5994 & 1.1633 \\
    Multi-MAML & GT & 16.2894 & 0.4330 & 12.3742 & 0.7791 \\ \hline
    \OurMethod (ours) & Softmax & 3.9140 & 0.4795 & 0.6889 & 0.4884 \\
    \OurMethod (ours) & Sigmoid & 1.4992 & 0.3414 & 2.4047 & 0.4414 \\
    \OurMethod (ours) & FiLM & 1.7094 & \textbf{0.3125} & 1.9234 & \textbf{0.4048}\\
    \end{tabular}}
    \label{tab:regression}
\end{table}

\vspaceAfterSection
\subsection{Regression}
\label{sec:regression}
\vspaceAfterSection
\begin{figure}[t]
	\centering
	\small
	\begin{tabular}{cccc}
	    Sinusoidal Functions & Linear Functions & Quadratic Functions &
	    \multirow{4}{*}{\includegraphics[width=0.25\textwidth, height=0.26\textwidth]{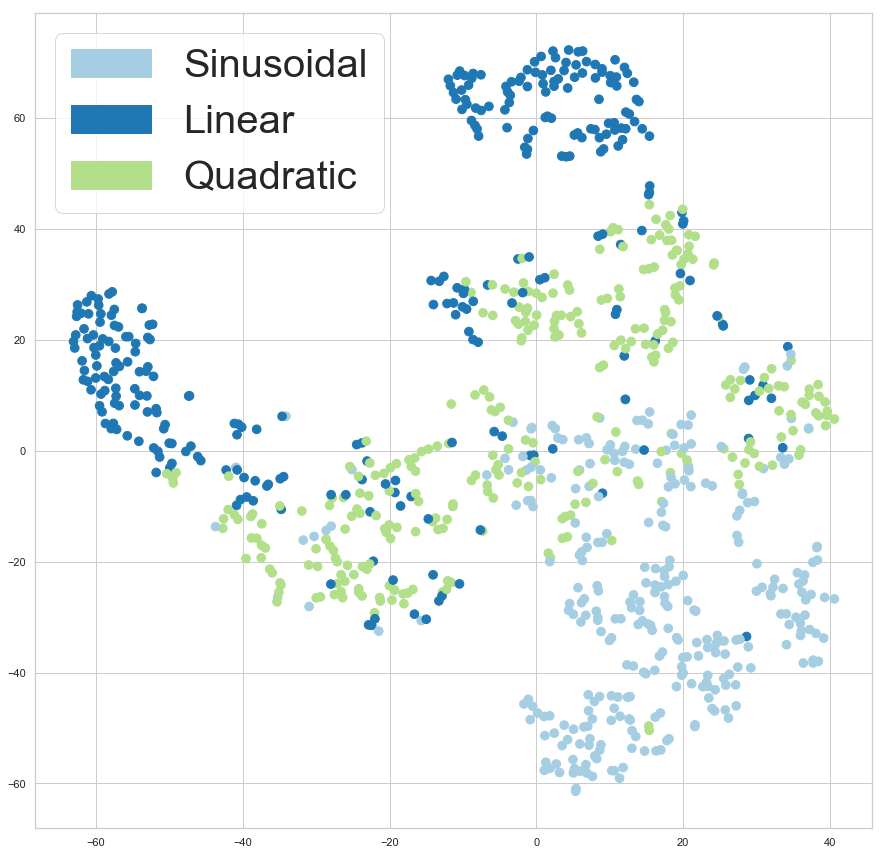}}\\
		\includegraphics[width=0.21\textwidth,trim={0.8cm 0 0.2cm 0},clip]{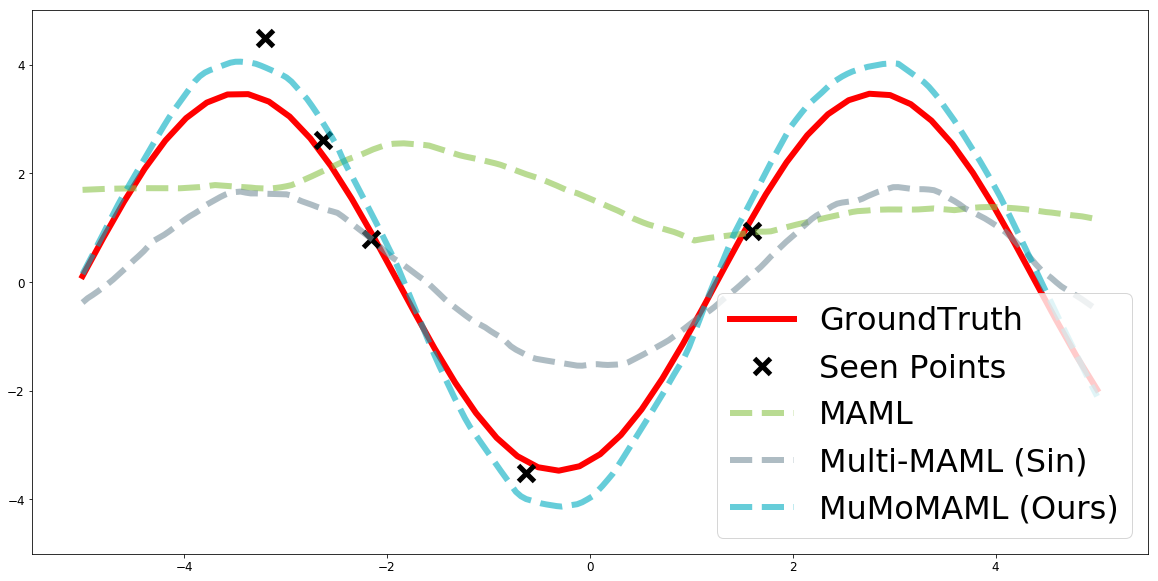} &
		\includegraphics[width=0.21\textwidth,trim={0.8cm 0 0.2cm 0},clip]{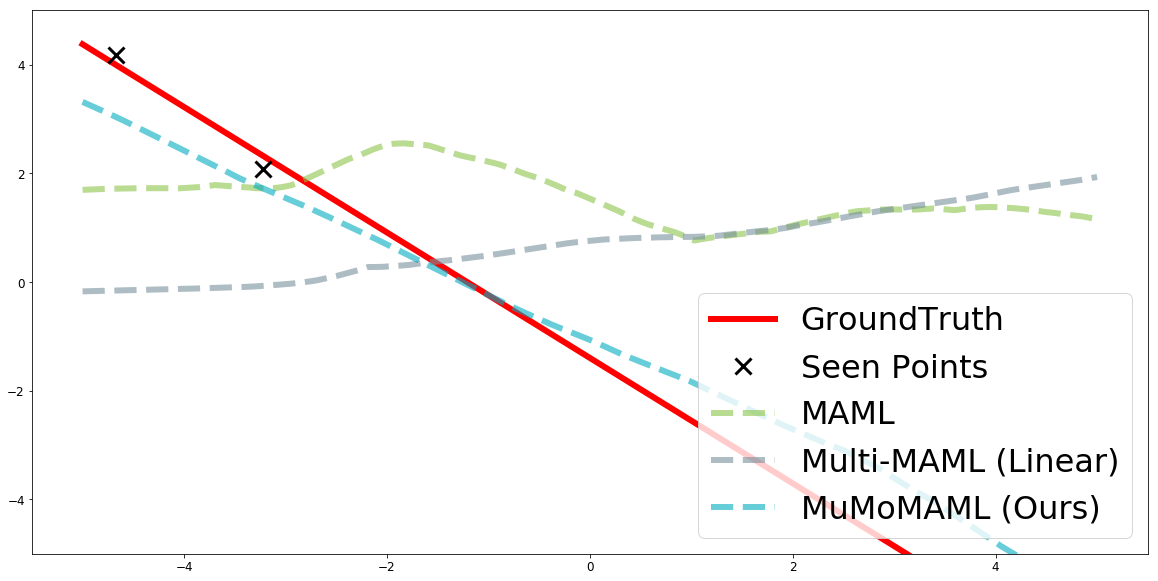} &
		\includegraphics[width=0.21\textwidth,trim={0.8cm 0 0.2cm 0},clip]{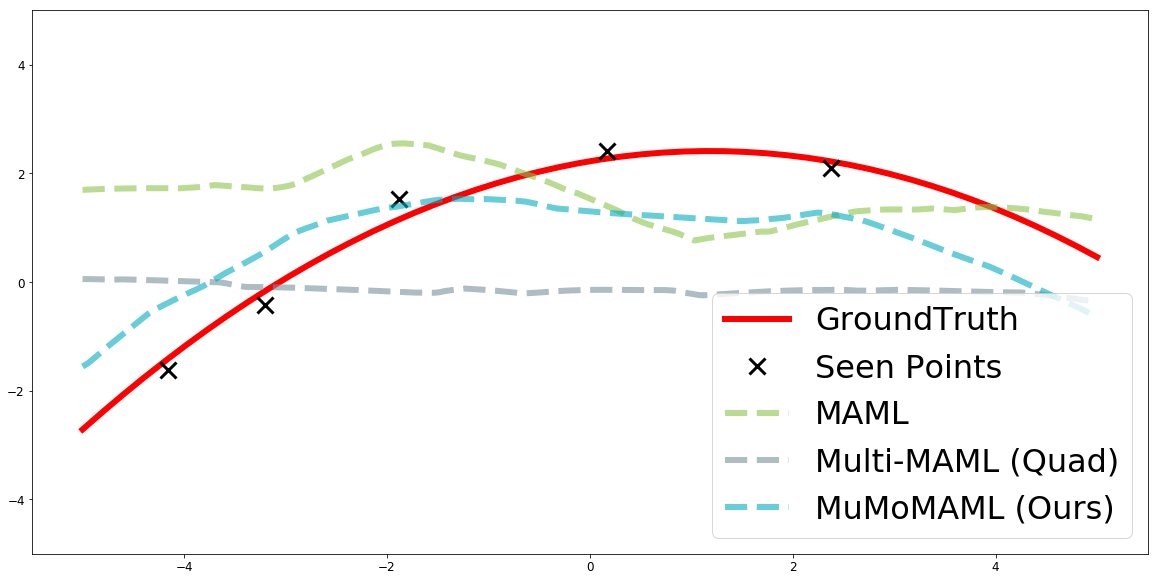} \\
		\multicolumn{3}{c}{(a) \OurMethod \textbf{after modulation} vs. other prior models}\\
		\includegraphics[width=0.21\textwidth,trim={0.8cm 0 0.2cm 0},clip]{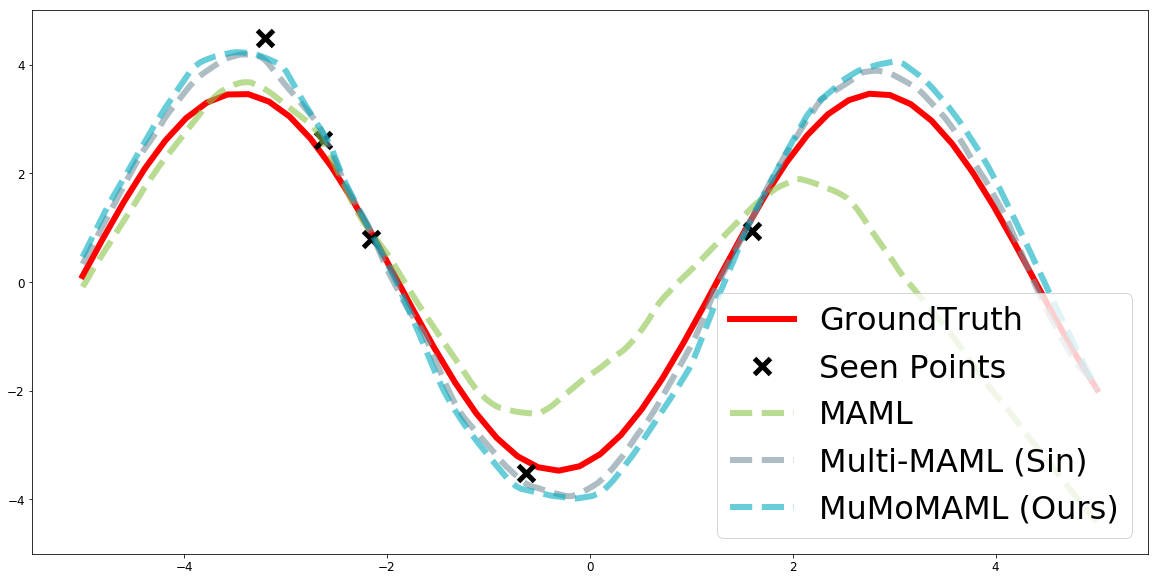} &
		\includegraphics[width=0.21\textwidth,trim={0.8cm 0 0.2cm 0},clip]{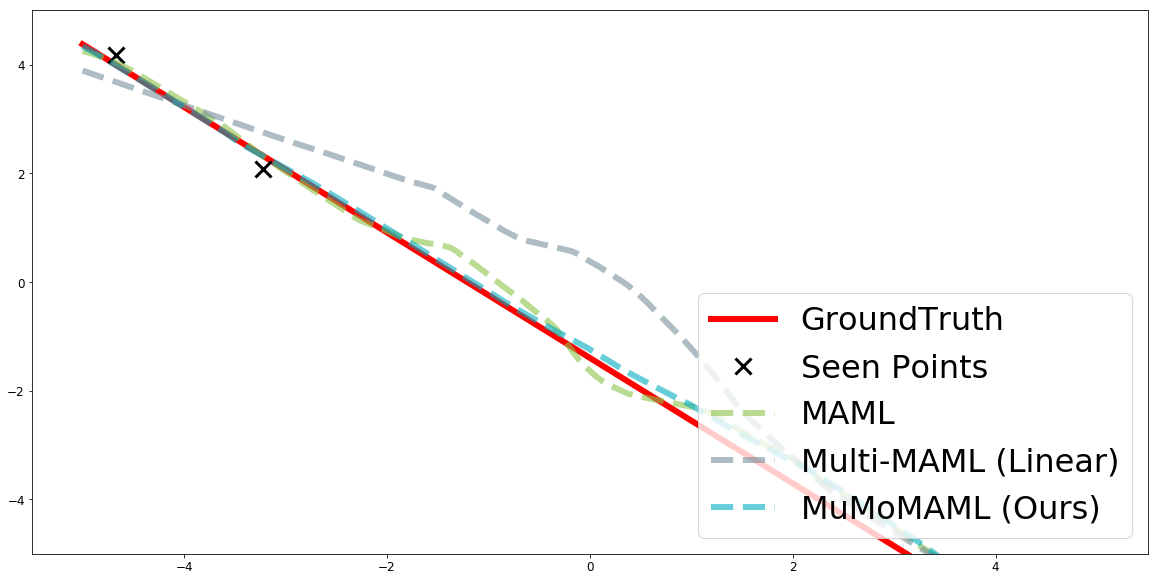} & \includegraphics[width=0.21\textwidth,trim={0.8cm 0 0.2cm 0},clip]{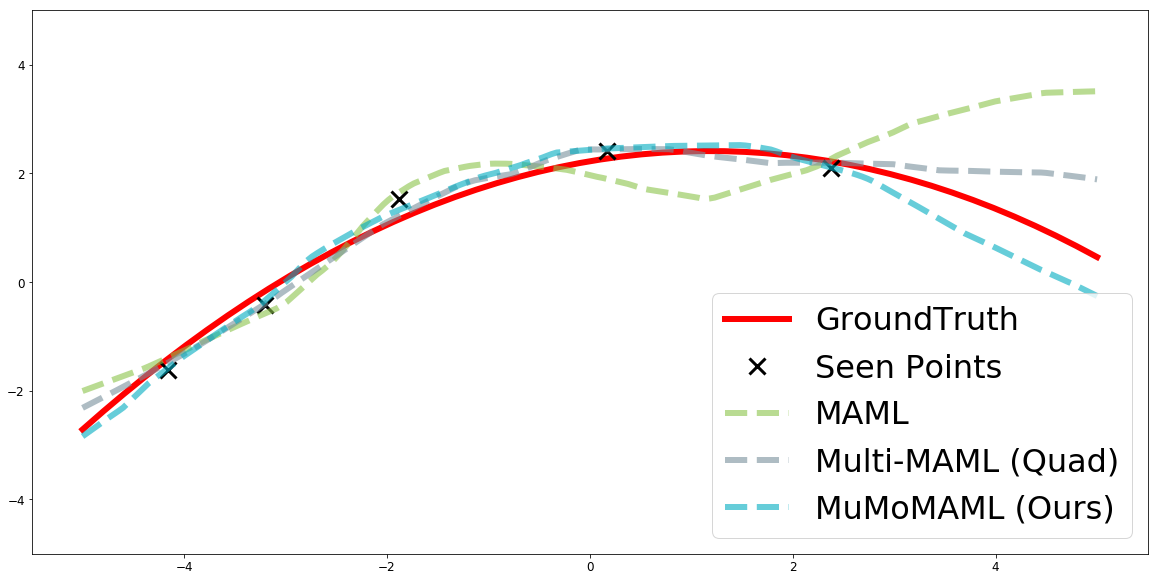} \\
		\multicolumn{3}{c}{(b) \OurMethod \textbf{after adaptation} vs. other posterior models} 
		& (c) Task embeddings \\
	\end{tabular}
	\caption{
    	\small
    	Few-shot adaptation for the multimodal regression task.
	    (a): Without any gradient update, \OurMethod (blue) fits target functions by modulating the meta-learned prior, outperforming the prior models of MAML (green) and Multi-MAML (gray). 
	    (b): After five steps of gradient updates, \OurMethod outperforms MAML and Multi-MAML on all functions. More visualizations in \myfig{fig:regression_A} and \myfig{fig:regression_B}.
	    (c): tSNE plots of the task embeddings $\upsilon$ produced by our model from randomly sampled tasks; marker color indicates different types of functions.
        The plot reveals a clear clustering according to different task modes, 
        showing that \OurMethod is able to infer the mode from a few samples and produce a meaningful embedding.
        The distance among distributions aligns with the intuition of the similarity of functions (\eg a quadratic function can sometimes be similar to a sinusoidal or a linear function while a sinusoidal function is usually different from a linear function).
	    \label{fig:regression}
	}
\end{figure}

We investigate our model's capability of learning on few-shot regression tasks
sampled from multimodal task distributions. In these tasks, a few input/output pairs
$\{x_k, y_k\}_{k=1,...,K}$ sampled from a one dimensional function are given 
and the model is asked to predict $L$ output values $y_1^q, ..., y_L^q$
for input queries $x_1^q, ..., x_L^q$.
We set up two regression settings with two task modes
(sinusoidal and linear functions) or three modes (quadratic functions added).
Please see \mysec{sec:experimental_details} for details.

As a baseline beside MAML, we propose Multi-MAML, which consists of $M$
(the number of modes) separate MAML models which are chosen for each query based
on ground-truth task-mode labels.
This baseline serves as an upper-bound for the performance of MAML
when the task-mode labels are available. 
The quantitative results are shown in \mytable{tab:regression}.
We observe that Multi-MAML outperforms MAML, showing that MAML's performance degrades on
multimodal task distributions. 
\OurMethod consistently achieves better results than Multi-MAML, demonstrating
that our model is able to discover and exploit transferable
knowledge across the modes to improve its performance.
The marginal gap between the performance of our model in two and three mode settings
indicates that \OurMethod is able to clearly identify the task modes and has sufficient
capacity for all modes.

We compared attention modulation with Sigmoid or Softmax and FiLM modulation and found that FiLM achieves better results. 
We therefore use FiLM for further experiments. Please refer to \mysec{sec:modulation} for additional details.
Qualitative results visualizing the predicted functions are shown in \myfig{fig:regression}.
\myfig{fig:regression} (a) shows that our model is able to identify tasks 
and fit to the sampled function well without performing gradient steps.
\myfig{fig:regression} (b) shows that our model consistently outperforms
the baselines with gradient updates.
\myfig{fig:regression} (c) plots a tSNE~\cite{maaten2008visualizing}, 
showing the model-based module is able to identify the task modes and produce embedding vectors $\upsilon$.
Additional results are shown in \mysec{sec:additional_results}.

\Skip{
We observe that \OurMethod is able to effectively modulate its meta-learned prior to fit a sampled task (see \myfig{fig:regression_result} (a)), which greatly eases the optimization procedure for our gradient-based learner to adapt (see \myfig{fig:regression_result} (b)).
To gain some insights of the task embeddings produced by our model, we perform tSNE~\cite{maaten2008visualizing} visualization on the predicted embedding vectors $\upsilon$ as \myfig{fig:tSNE} (a). 
It shows that our model is able to capture the mode structure in the embeddings, allowing the performance gain from modulation.
}

\vspaceAfterSection
\subsection{Reinforcement Learning}
\label{sec:RL}
\vspaceAfterSection



\begin{figure*}[t]
    \centering
    \begin{tabular}{cccc}
    \includegraphics[width=0.21\textwidth]{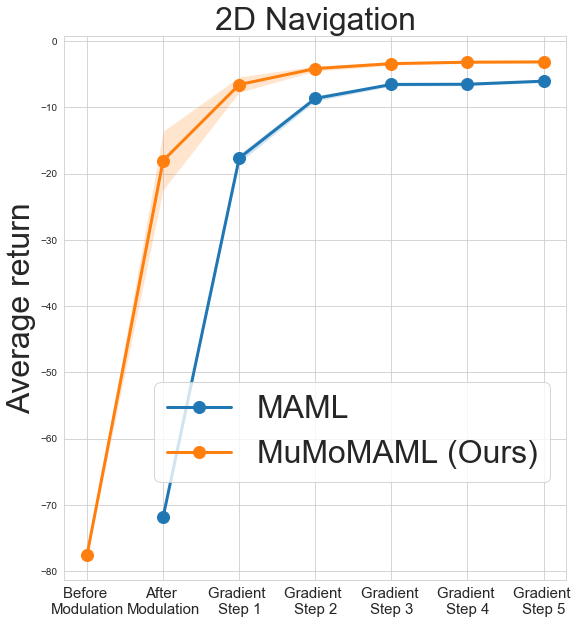} &
    \includegraphics[width=0.21\textwidth]{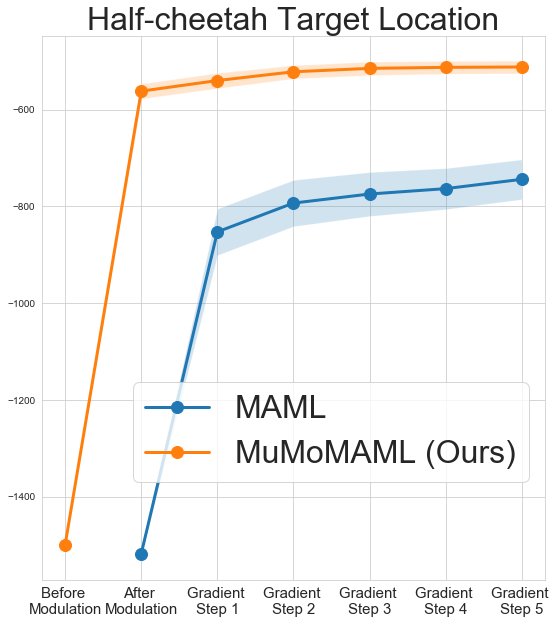} &
    \includegraphics[width=0.21\textwidth]{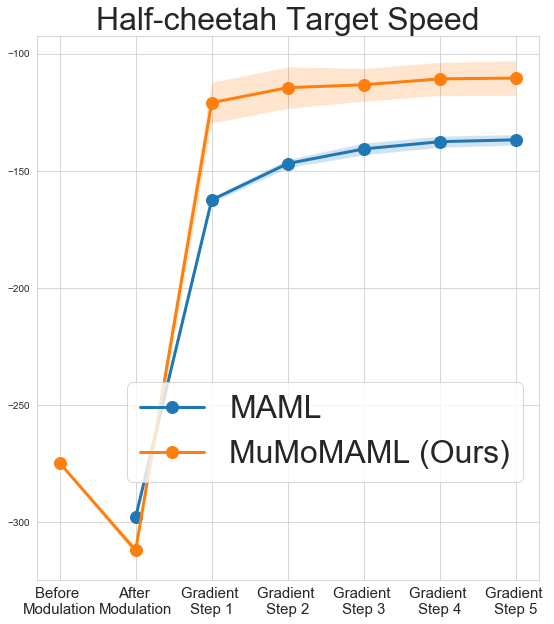} &
    \includegraphics[width=0.24\textwidth]{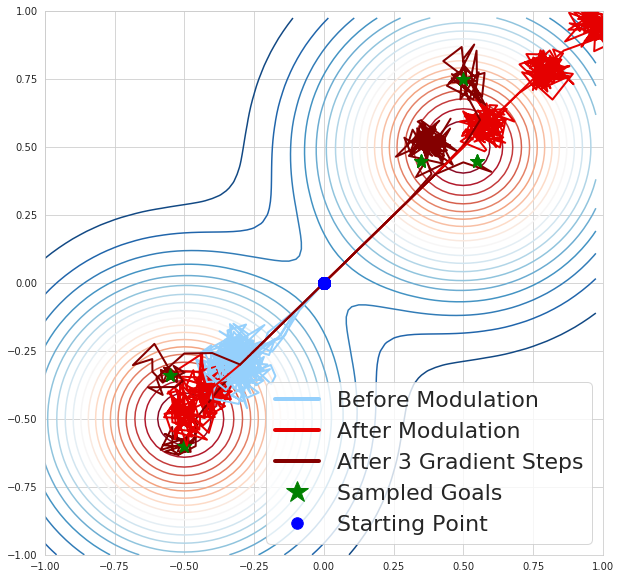}
    \\
    \small{(a) 2D Navigation} & \small{(b) Target Location} & \small{(c) Target Speed} & \small{(d) 2D Navigation Results}
    
    \end{tabular}
    \caption{ \small
        (a-c) Adaptation curves for \OurMethod and MAML baseline in 2D navigation and half-cheetah environments. The ``after modulation'' step represents the rewards of the modulated policy for \OurMethod and the initial rewards for MAML. \OurMethod outperforms MAML across the gradient update steps given a single extra trajectory.
        (d) Visualized trajectories sampled using \OurMethod in the 2D navigation environment. The contours represent the probability density of the goal distribution (red: high probability; blue: low probability). The trajectories demonstrate the effect of modulation and the subsequent fine tuning with gradient steps. Additional trajectory visualizations can be found in \myfig{fig:rl_extra_results}.
        \label{fig:adaptation_figures}
    }
\end{figure*}

\begin{wrapfigure}{r}{0.29\textwidth}

    \centering
    \includegraphics[width=0.27\textwidth, height=0.25\textwidth ]{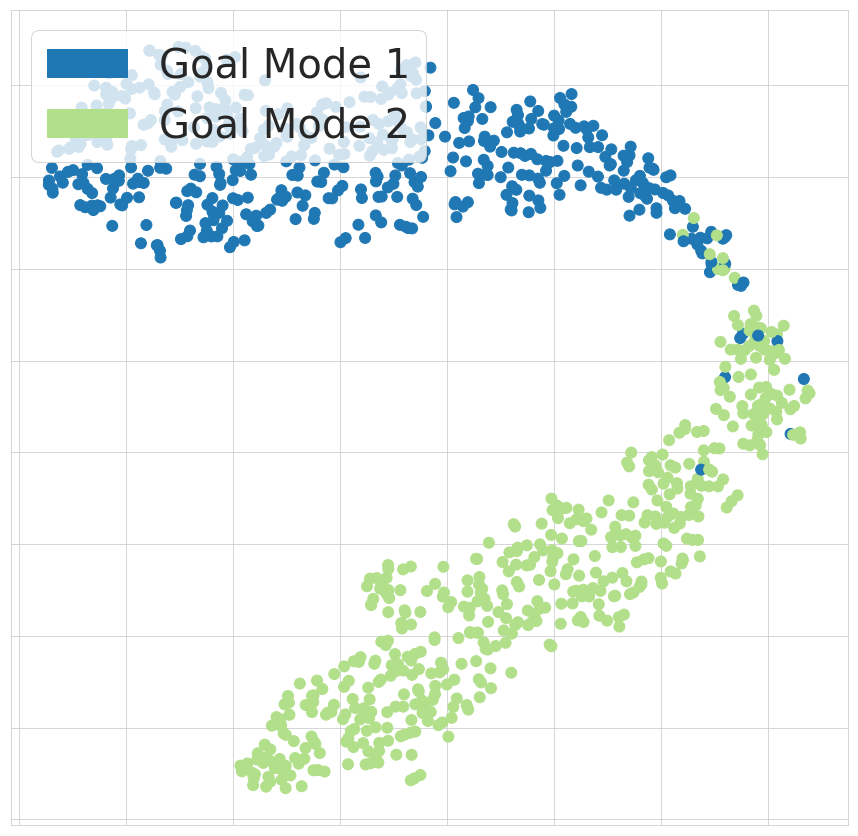}
    
    \caption{ \small
        A tSNE plot of task embeddings of randomly sampled tasks in Target Location environment capturing the bimodal task distribution.
        \label{fig:rl_embedding}
    }
\end{wrapfigure}

We experiment with \OurMethod in three reinforcement learning (RL) environments
to verify its ability to learn to rapidly adapt to tasks sampled from multimodal task distributions
given a minimum amount of interaction with an environment.
\footnote{Please refer to \mysec{sec:experimental_details} for details on the experimental setting.}

\noindent \textbf{2D Navigation.} 
We utilize a 2D navigation environment with bimodal task distribution to investigate the capabilities of
the embedding network to identify the task mode based on trajectories sampled from RL environments and 
the modulation network to modulate a policy network.
In this environment, the agent is rewarded for moving close to a goal location.
The model-based meta-learner is able to 
identify the task modes and modulate the policy accordingly,
allowing efficient fast adaptation.
This is shown in the agent trajectories and the average return plots presented in \myfig{fig:adaptation_figures} (a) and (d),
where our model outperforms MAML with any number of gradient steps.

\noindent \textbf{Half-cheetah Target Location and Speed.} 
To investigate the scalability of our method to more complex RL environments we experiment with 
locomotion tasks based on the half-cheetah model. In the target location and target
speed environments the agent is rewarded for moving close to the target location or
moving at target speed respectively. The targets are sampled from bimodal distributions.
In these environments, the dynamics are considerably more complex than in
the 2D navigation case. 
\OurMethod is able to utilize the model-based meta-learner
to effectively modulate the policy network and retain an advantage over MAML across all
gradient update steps as seen from the adaptation curves in
\myfig{fig:adaptation_figures} (b) and \myfig{fig:adaptation_figures} (c).
A tSNE plot of the embeddings in \myfig{fig:rl_embedding} shows that our model is able to produce meaningful task embeddings $\upsilon$.


\vspaceAfterSection
\section{Conclusion}
\label{sec:conclusion}
\vspaceAfterSection
We presented a novel meta-learning approach that is able to leverage the strengths of both model-based and gradient-based meta-learners to discover and exploit the structure of multimodal task distributions. 
With the ability to effectively recognize the task modes as well as rapidly adapt through a few gradient steps,
our proposed \OurMethod achieved superior generalization performance on multimodal few-shot regression, reinforcement learning, and image classification.


\clearpage

\appendix

\section{Modulation Methods}
\label{sec:modulation}

To allow efficient adaptation, the modulation network activates or deactivates units
of each network block of the gradient-based meta-learner according to the given target
task embedding. 
We investigated a representative set of modulation operations,
including attention-based modulation~\citep{mnih2014recurrent,vaswani_attention_2017} 
and feature-wise linear modulation (FiLM)~\citep{perez_film:_2017}. 

\paragraph{Attention based modulation} ~\citep{mnih2014recurrent,vaswani_attention_2017} has been widely used in modern deep learning models and has proved its effectiveness across various tasks~\citep{yang2016stacked,mnih2014recurrent,zhang_self-attention_2018,xu2015show}.
Inspired by the previous works, we employed attention to modulate the prior model. In concrete terms, attention over the outputs of all neurons (Softmax) or a binary gating value (Sigmoid) on each neuron's output is computed by the model-based meta-learner. These parameters $\tau$ are then used to scale the pre-activation of each neural network layer $\mathbf{F}_{\theta}$, such that $\mathbf{F}_{\phi} = \mathbf{F}_{\theta} \otimes \tau$. Note that here $\otimes$ represents a channel-wise multiplication. 
 
\paragraph{Feature-wise linear modulation (FiLM)}~\citep{perez_film:_2017} proposed to modulate neural networks to condition the networks on data from different modalities. We adopt FiLM as an option for modulating our gradient-based meta-learner. Specifically, the parameters $\tau$ are divided in to two components $\tau_{\gamma}$ and $\tau_{\beta}$ such that for a certain layer of the neural network with its pre-activation $\mathbf{F}_{\theta}$, we would have $\mathbf{F}_{\phi} = \mathbf{F}_{\theta} \otimes \tau_{\gamma} + \tau_{\beta}$. It can be viewed as a more generic form of attention mechanism. Please refer to~\cite{perez_film:_2017} for the complete details. 

As shown in the quantitative results (\mytable{tab:regression}),
using FiLM as a modulation method achieves 
better results comparing to attention mechanism with Sigmoid or Softmax. 
We therefore use FiLM for further experiments.

\section{Few-shot Image Classification}
\label{sec:few-shot}

\begin{figure*}[t]
    \centering
    \begin{tabular}{cccc}
    \includegraphics[width=0.21\textwidth]{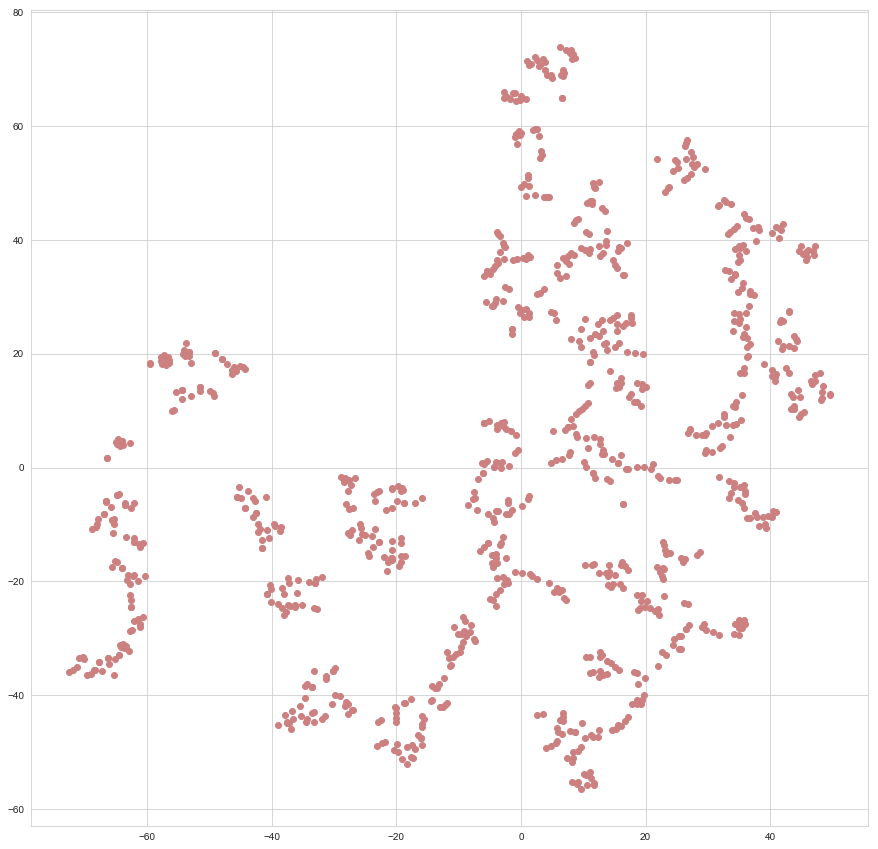} &
    \includegraphics[width=0.21\textwidth]{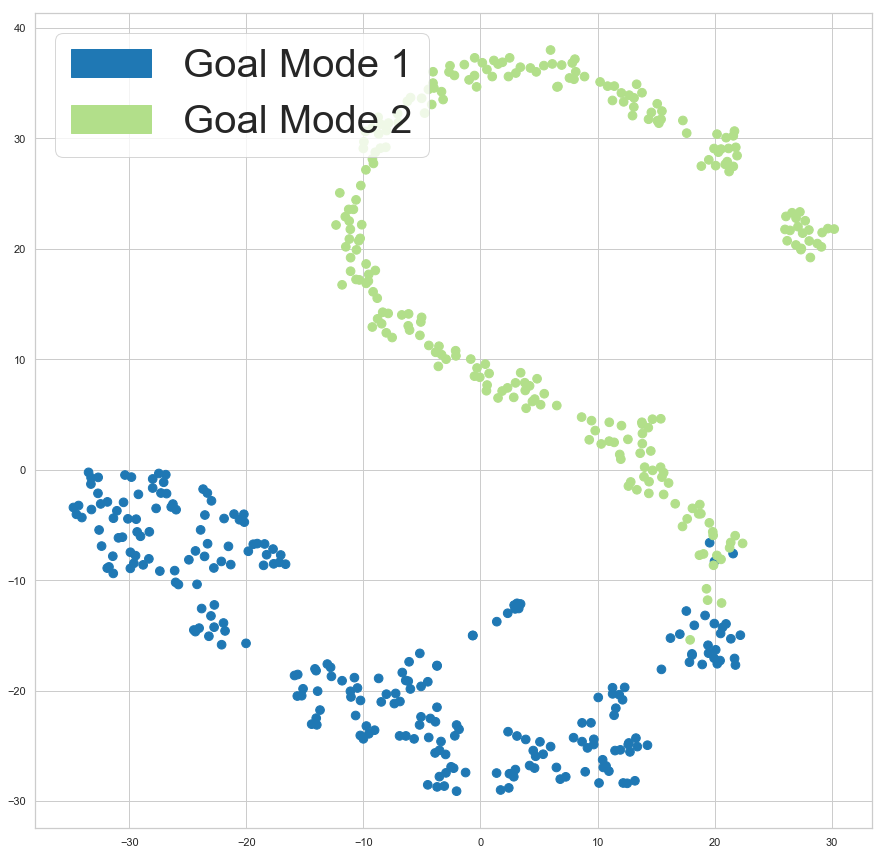} &
    \includegraphics[width=0.21\textwidth]{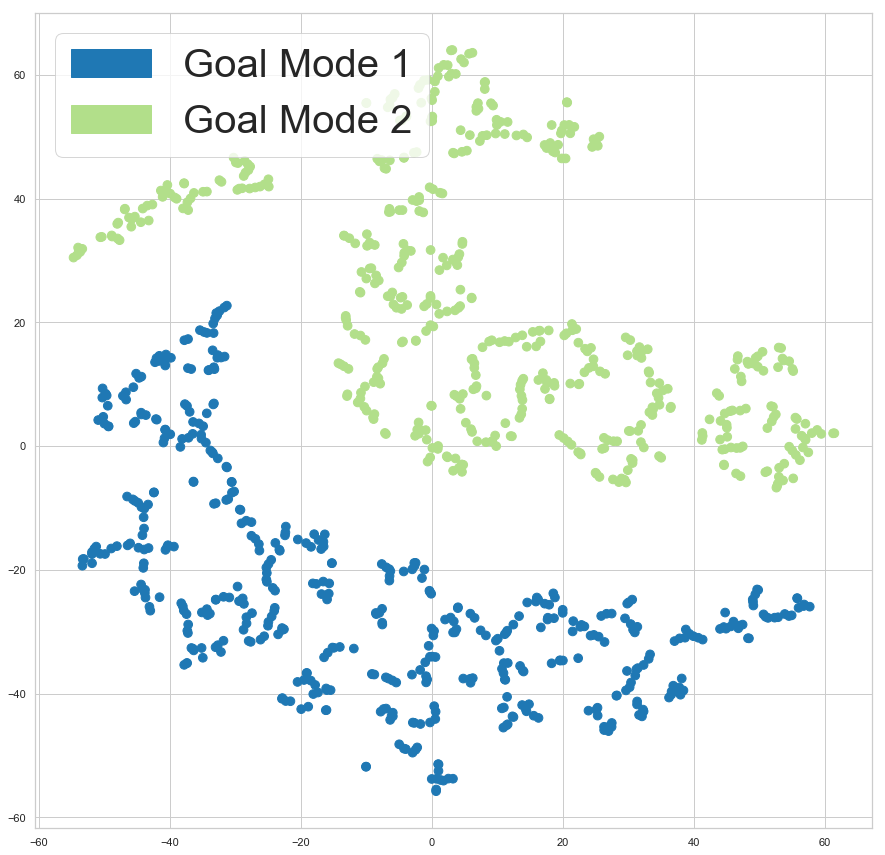} &
    \includegraphics[width=0.21\textwidth]{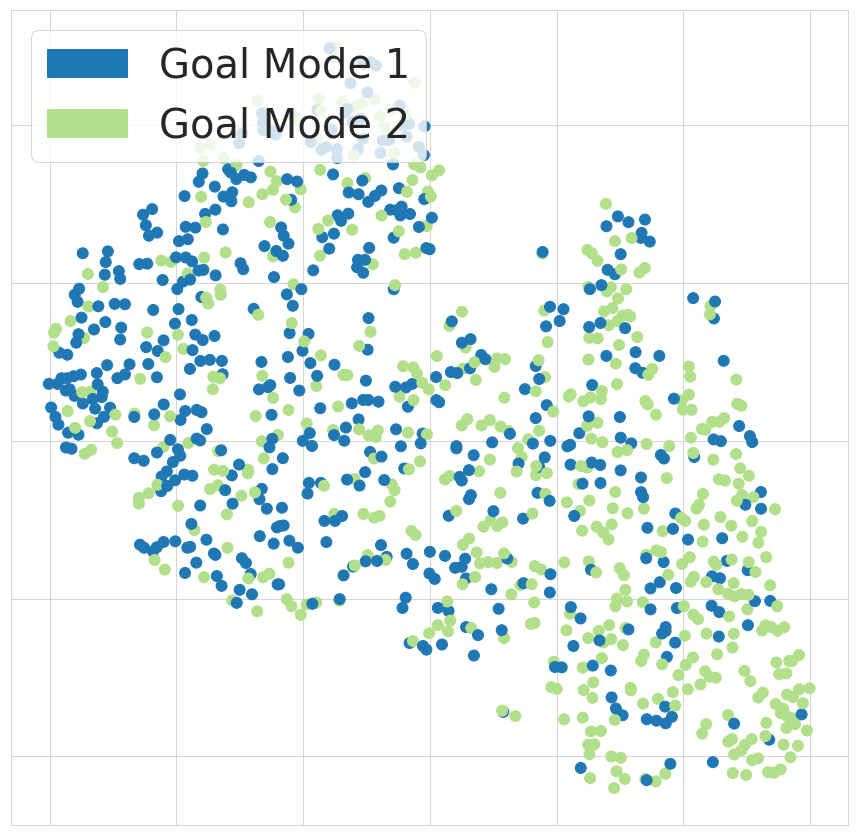} \\
    \small{(a) Few-shot} & \small{(b) 2D Navigation} & \small{(c) 2D Navigation} & \small{(d) Target Speed} \\
    \small{classification} & \small{interpolated goals} & \small{random goals} & \small{random goals} \\

    \end{tabular}
    \caption{ \small
        tSNE plots of task embeddings produced in different problem settings. (a) Embeddings of randomly sampled tasks in few-shot image classification. (b-c) Embeddings of tasks with goal locations interpolated between the modes and sampled randomly in 2D Navigation. (d) Randomly sampled goals in target speed environment. 
        \label{fig:few-shot_embedding}
    }
\end{figure*}

The task of few-shot image classification considers a problem of
classifying images into N classes with a small number (K) of labeled samples available. 
To evaluate our model on this task, we conduct experiments on
\textsc{Omniglot}, a widely used handwritten character dataset of binary images. 
The results are shown in \mytable{table:few-shot}, demonstrating that
our method achieves comparable or better results against state-of-the-art algorithms.

To gain insights to the task embeddings $\upsilon$ produced by our model,
we sampled 2000 tasks randomly and employ tSNE to visualize the $\upsilon$ in \myfig{fig:few-shot_embedding} (a).
While we are not able to clearly distinguish the modes of task distributions, we observe that the distribution of the produced embeddings is not uniformly distributed or unimodal, potentially indicating the multimodal nature of this task.

\begin{table}[ht]
\centering
    \caption{ \small
        5-way and 20-way, 1-shot and 5-shot classification accuracy on \textsc{Omniglot} Dataset.
        For each task, the best-performing method is highlighted.
        \OurMethod achieves comparable or better results against state-of-the-art few-shot learning algorithms for image classification. 
    }
    \scalebox{0.85}{\begin{tabular}{c|cc|cc} \\
    \multirow{3}{*}{Method} & \multicolumn{4}{c}{\textsc{Omniglot}} \\
     & \multicolumn{2}{c}{\textbf{5 Way Accuracy (in \%)}} & \multicolumn{2}{c}{\textbf{20 Way Accuracy (in \%)}} \\ 
    & 1-shot & 5-shot & 1-shot & 5-shot \\ 
    \hlineB{4}
    Siamese nets~\cite{koch2015siamese} & 97.3 & 98.4 & 88.2 & 97.0 \\
    Matching nets~\cite{vinyals2016matching} &  98.1 & 98.9 & 93.8 & 98.5  \\
    Meta-SGD ~\cite{li_meta-sgd:_2017}  & 99.5 & \textbf{99.9} &  95.9 & 99.0 \\
    Prototypical nets~\cite{snell_prototypical_2017} & 97.4 & 99.3 & 96.0 & 98.9 \\
    SNAIL~\cite{mishra_simple_2017} &  99.1 & 99.8 & \textbf{97.6} & \textbf{99.4} \\
    T-net~\cite{lee_gradient-based_2018}  & 99.4 & - & 96.1 & - \\
    MT-net~\cite{lee_gradient-based_2018} & 99.5 & - & 96.2 & - \\ \hline
    MAML~\citep{finn_model-agnostic_2017} & 98.7 & \textbf{99.9} &  95.8 & 98.9 \\
    \OurMethod (ours) & \textbf{99.7} & \textbf{99.9} & 97.2 & \textbf{99.4} 
    
    \label{table:few-shot}
    \end{tabular}}
\end{table}


\section{Implementation Details}
\label{subsec:implementation}

For the model-based meta-learner, we used \SeqToSeq~\citep{sutskever2014sequence} encoder structure to encode the sequence of $\{x, y\}_{k=1,...,K}$ with a bidirectional GRU~\citep{chung2014empirical} and use the last hidden state of the recurrent model as the representation for the task. We then apply a one-hidden-layer multi-layer perception (MLP) for each layer in the gradient-based learner's model to generate the set of task-specific parameters $\tau_i$, as described in the previous section. We implemented our models for three representative learning scenarios -- regression, few-shot learning and reinforcement learning. The concrete architecture for each task might be different due to each task's data format and nature. We discuss them in the section~\ref{sec:experiments}.

\section{Additional Experimental Details}
\label{sec:experimental_details}
\subsection{Regression}

\paragraph{Setups.} To form multimodal task distributions, we consider a family of functions including
sinusoidal functions (in forms of $A \cdot \sin{w \cdot x + b} + \epsilon$, with $A \in [0.1, 5.0]$, $w \in [0.5, 2.0]$ and $b \in [0, 2\pi ]$), 
linear functions (in forms of $A \cdot x + b$, with $A \in [-3, 3]$ and $b \in [-3, 3] $)
and quadratic functions (in forms of $A \cdot (x - c)^2 + b$, with $A \in [-0.15, -0.02] \cup [0.02, 0.15]$, $c \in [-3.0, 3.0]$ and $b \in [-3.0, 3.0]$ ). Gaussian observation noise with $\mu=0$ and $\epsilon=0.3$ is added to each data point sampled from the target task. In all the experiments, $K$ is set to $5$ and $L$ is set to $10$. We report the mean squared error (MSE) as the evaluation criterion. 
Due to the multimodality and uncertainty, this setting is more challenging comparing to~\citep{finn_model-agnostic_2017}. 

\paragraph{Models and Optimization.} In the regression task, we trained a 4-layer fully connected neural network with the hidden dimensions of $100$ and ReLU non-linearity for each layer, as the base model for both MAML and \OurMethod. In \OurMethod, an additional model with a Bidirectional GRU of hidden size $40$ is trained to generate $\tau$ and to modulate each layer of the base model. We used the same hyper-parameter settings as the regression experiments presented in ~\cite{finn_model-agnostic_2017} and used Adam~\cite{kingma2014adam} as the meta-optimizer. For all our models, we train on 5 meta-train examples and evaluate on 10 meta-val examples to compute the loss.

\subsection{Reinforcement Learning}
\label{subsec:rl_appendix}

\begin{figure*}[t]
    \centering
    \begin{tabular}{ccc}
    \includegraphics[width=0.3\textwidth]{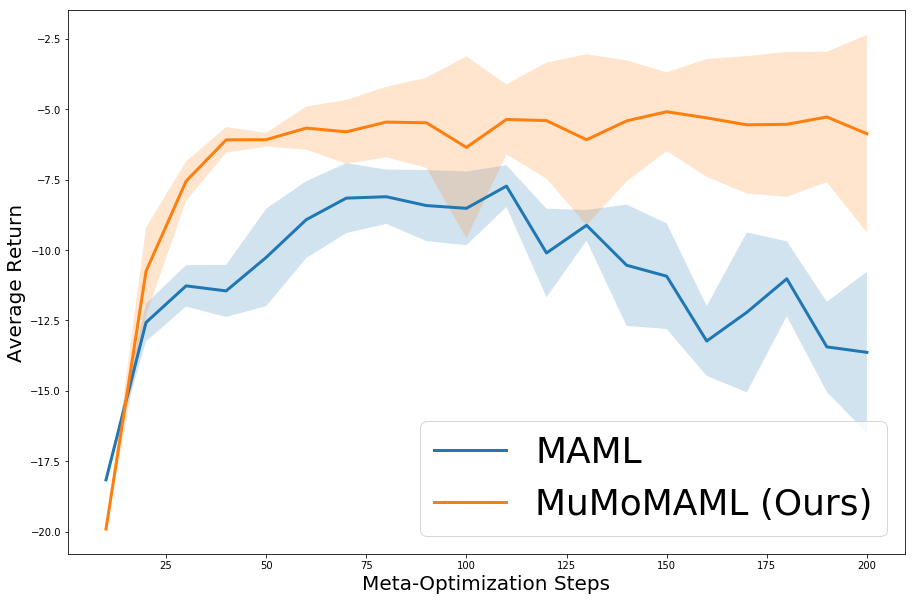} &
    \includegraphics[width=0.3\textwidth]{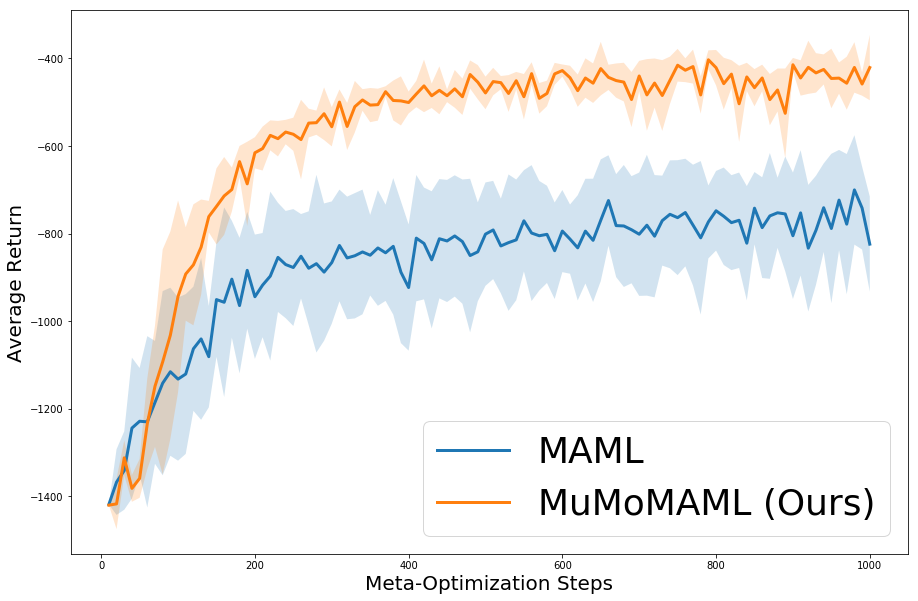} &
    \includegraphics[width=0.3\textwidth]{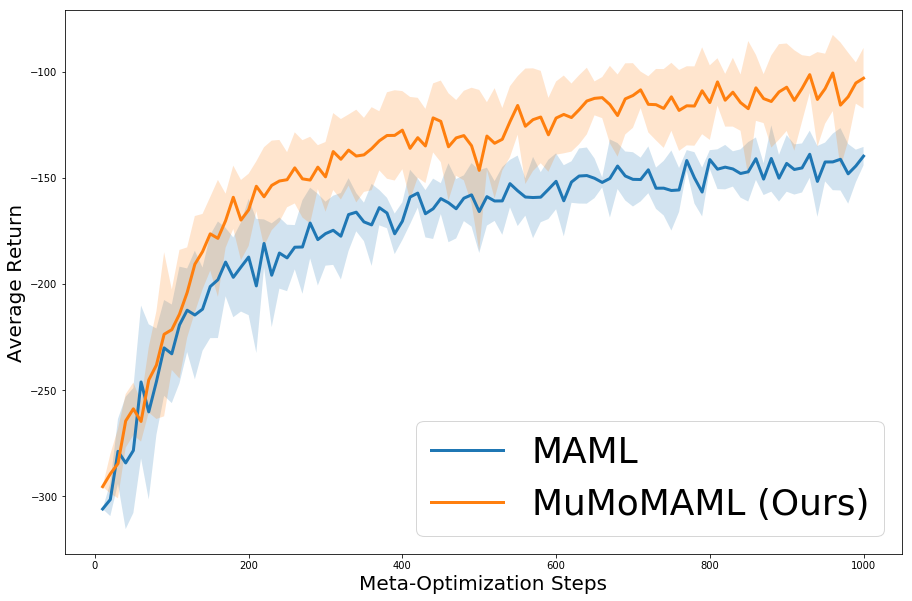} \\
    \small{(a) 2D Navigation} & \small{(b) Target Location} & \small{(c) Target Speed} \\
    
    \end{tabular}
    \caption{
    \small
    Training curves of \OurMethod and MAML in reinforcement learning environments.
    The value indicates the performance evaluated after modulation (our model) and 5 gradient steps.
    The plots demonstrate that our model consistently outperforms MAML from the beginning of training to the end.
    }
    \label{fig:rl_training_curves}
\end{figure*}

Along with few-shot classification and regression, reinforcement learning has been a
central problem where meta-learning has been studied \cite{schmidhuber1997shifting,schmidhuber1998reinforcement,wang2016learning,finn_model-agnostic_2017,mishra_simple_2017}.

To identify the mode of a task distribution in the context of reinforcement learning,
we run the meta-learned prior model without modulation or adaptation 
to interact with the environment and collect a single trajectory and obtained rewards.
Then the collected trajectory and rewards are fed to
our model-based meta-learner to compute the task embedding $\upsilon$ and $\tau$.
With this minimal amount of interaction with the environment, our model is
able to recognize the tasks and modulate the policy network to effectively learn
in multimodal task distributions.

The batch of trajectories used for computing the first gradient-based adaptation
step is sampled using the modulated model and the batches after that using
the modulated and adapted model from the previous update step.
We follow the MAML gradient-based adaptation procedure. For the gradient
adaptation steps, we use the vanilla policy gradient algorithm
\cite{williams1992simple}. As the meta-optimizer we use the trust
region policy optimization algorithm \cite{schulman_trust_2015}.

\paragraph{Models and Optimization.} We use embedding model hidden size of 128 and modulation
network hidden size of 32 for all environments.  
The training curves for all environments are presented in \myfig{fig:rl_training_curves},
which show that our proposed model consistently
outperforms MAML from the beginning of training to the end.
During optimization we save model parameters and evaluate the model with five 
gradient update steps every 10 meta update steps. 
We compute the adaptation curves presented in \myfig{fig:adaptation_figures}
using the model which achieved the best score after the five gradient updates during
training. 

\paragraph{2D-Navigation}
In the 2D navigation environment the goals are sampled with equal probability
from one of two bivariate Gaussians with means of $(0.5, 0.5)$ and $(-0.5, -0.5)$
and standard deviation of $0.1$.
In the beginning of each episode, the agent starts at the origin.
The agent's observation is its 2D-location and
the reward is the negative distance to the goal. The agent does
not observe the goal directly, instead it must learn to navigate there based on
the reward function alone. The agent outputs vectors which elements are clipped to
the range $[-0.1, 0.1]$ and the agent is moved in the environment by the clipped vector.
The episode terminates after 100 steps or
when the agent comes to the distance of $0.01$ from the goal.

For both \OurMethod and MAML we sample 20 trajectories for computing the
gradient-based adaptation steps and 20 tasks for meta update steps.
We use inner loop update step size of $0.1$ for MAML and $0.05$ for \OurMethod.
We train both methods for 200 meta-optimization steps. 

We investigate the behavior of the task embedding network by sampling tasks from the
environment and computing a tSNE plot of the task embeddings. A tSNE plot for
goals interpolated between the goal modes is presented in \myfig{fig:few-shot_embedding}
(b) and a plot for randomly sampled goals is presented in (c). The
tSNE plots show that the structure of the embedding space reflects the goal distribution.

\paragraph{Target Location}
Target location is a locomotion environment based on the half-cheetah model in the
mujoco \cite{todorov2012mujoco} simulation framework.
The environment design follows \cite{finn_model-agnostic_2017}, except for the
reward definition. 
The reward on each time step is
$$R(s) = -1 * abs(x_{torso} - x_{goal}) + \lambda_{control} * \|a\|^{2}$$
where $x_{torso}$ and $x_{goal}$ are the x-positions of the midpoint of the
half-cheetah's torso and the target location
respectively, $\lambda_{control} = -0.05$ is the coefficient for the control
penalty and $a$ is the action
chosen by the agent. The target location is sampled from a distribution
consisting of two Gaussians with means of $-7$ and $7$ and standard deviation
of $2$.
The observation is the location and movement state of the joints of the half-cheetah.
The episode terminates after 200 steps.

For both \OurMethod and MAML we sample 20 trajectories for computing the
gradient-based adaptation steps and 40 tasks for meta update steps.
We use inner loop update step size of $0.05$ for both methods.
Both methods are trained for 1000 meta-optimization steps.

\paragraph{Target Speed}
Target speed is another half-cheetah based locomotion environment.
The environment design is similar to the target location environment, except the
reward is based on achieving target speed.
The reward on each time step is
$$R(s) = -1 * abs(v_{agent} - v_{target}) + \lambda_{control} * \|a\|^{2}$$
where $v_{agent}$ and $v_{target}$ are the speed of the head of the
half-cheetah and the target speed
respectively, $\lambda_{control} = -0.05$ is the coefficient for the control
penalty and $a$ is the action
chosen by the agent. The target speed is sampled from a distribution
consisting of two Gaussians with means of $-1$ and $1.5$
and standard deviation of $0.5$.
The observation and other environment details are the same
as in the target location environment.
For training \OurMethod and MAML same hyperparameters are used in as for
target location environment.

A tSNE plot of randomly sampled task embeddings from the target speed
environment is presented in \myfig{fig:few-shot_embedding} (d). The
embeddings for tasks from different modes are distributed towards the
opposite ends of the tSNE plot, but the modes are not as clearly distinguishable
as in the other environments. Also, the modulated policy in the target speed environment
achieves lower returns than in other environments as is evident from 
\myfig{fig:adaptation_figures}. Notice that in the reinforcement
learning setting, the model is not optimized to achieve high returns immediately
after the modulation step but only after modulation and one gradient update. After
one gradient step \OurMethod consistently outperforms MAML in target speed as well.

\subsection{Few-shot Image Classification}

\paragraph{Setups.} In the few-shot learning experiments, we used \textsc{Omniglot}, a dataset consists of 50 languages, with a total of 1632 different classes with 20 instances per class. Following \cite{santoro_meta-learning_2016}, we downsampled the images to $28 \times 28$ and perform data augmentation by rotating each member of an existing class by a multiple of 90 degrees to form new data points.

\paragraph{Models and Optimization.} Following prior works~\citep{vinyals2016matching,finn_model-agnostic_2017}, we used the same 4-layer convolutional neural network and applied the same training and testing splits from \cite{finn_model-agnostic_2017} and compare our model against baselines for 5-way and 20-way, 1-shot and 5-shot classification.

\section{Additional Experimental Results}
\label{sec:additional_results}
\subsection{Additional Results for Regression}

\begin{figure*}[t]
\centering
    \begin{tabular}{cc}
    \includegraphics[width=.42\textwidth]{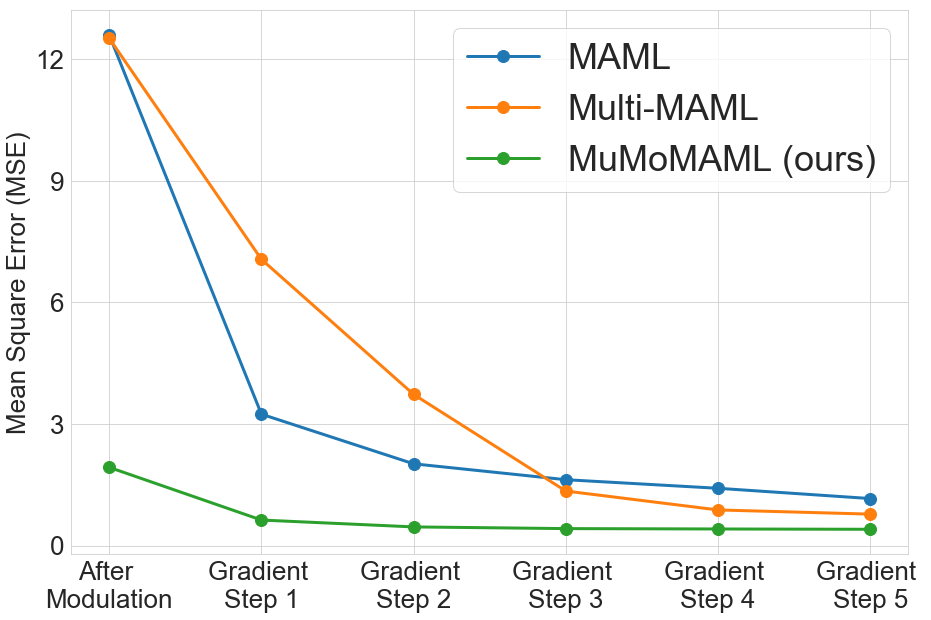} &
    \includegraphics[width=.385\textwidth]{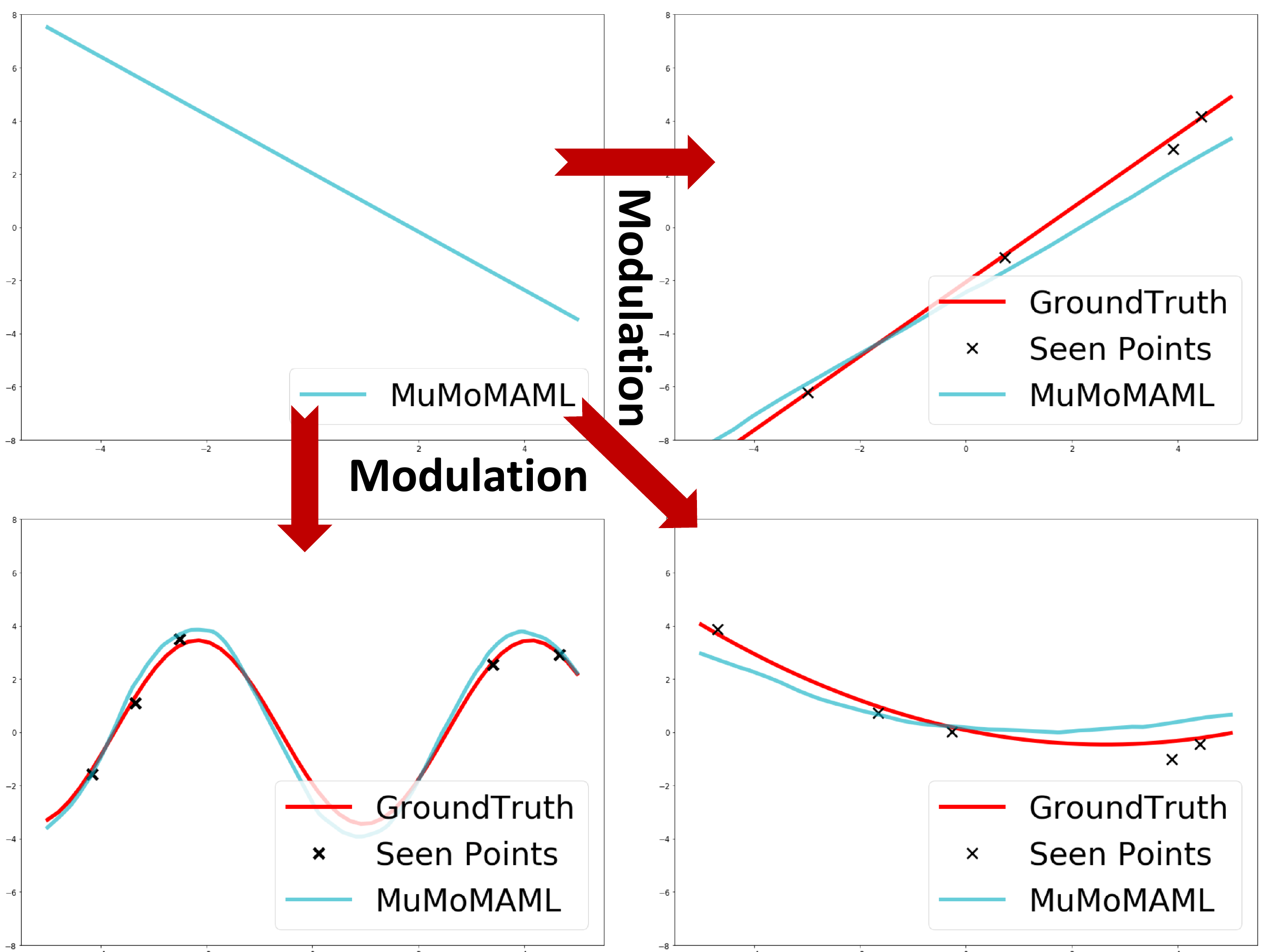}  \\
    \small{(a) Performance upon updates} & (b) \small{Effect of modulation}
    \end{tabular}
    
    \caption{
    \small
    \textbf{(a)} Comparing the models' performance with respect to the number of gradient updates applied. For \OurMethod, we report the performance after modulation for gradient step 0. 
    \textbf{(b)} A demonstration of the modulation on prior model by our model-based meta-learner. With the FiLM modulation, \OurMethod can adapt to different priors before gradient-based adaptation.
    }
    \label{fig:regression_result}
\end{figure*}

\myfig{fig:regression_result} demonstrates that \OurMethod outperforms
the MAML and Multi-MAML baselines no matter how many gradient steps are performed.
Also, \OurMethod is able to achieve good performance solely based on modulation without any gradient update,
showing that our model-based meta-learner is capable of identifying the mode of a multimodal task distribution and effectively modulate the meta-learned prior.

\begin{figure}[ht]
	\centering
	\small
	\begin{tabular}{ccc}
	     Sinusoidal Functions & Linear Functions & Quadratic Functions \\
		\includegraphics[width=0.305\textwidth,trim={0.8cm 0 0.2cm 0},clip]{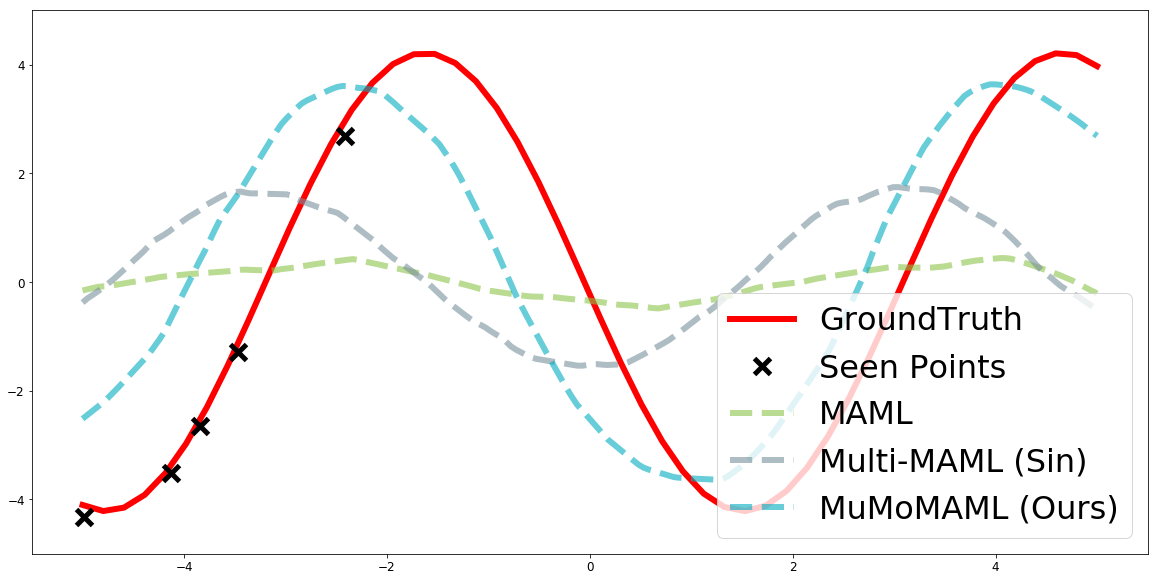} &
		\includegraphics[width=0.305\textwidth,trim={0.8cm 0 0.2cm 0},clip]{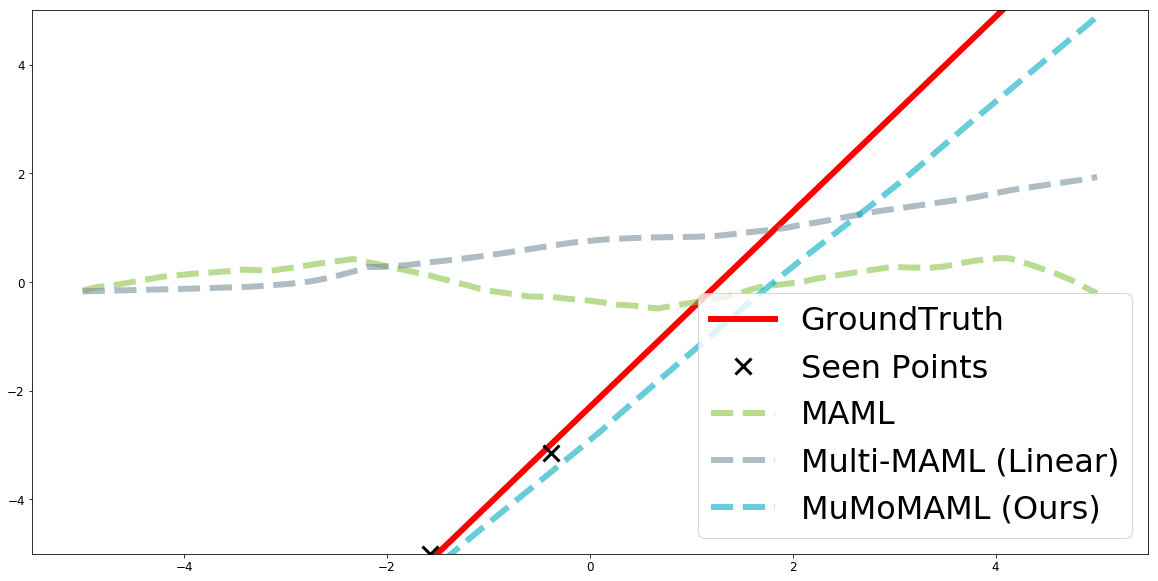} &
		\includegraphics[width=0.305\textwidth,trim={0.8cm 0 0.2cm 0},clip]{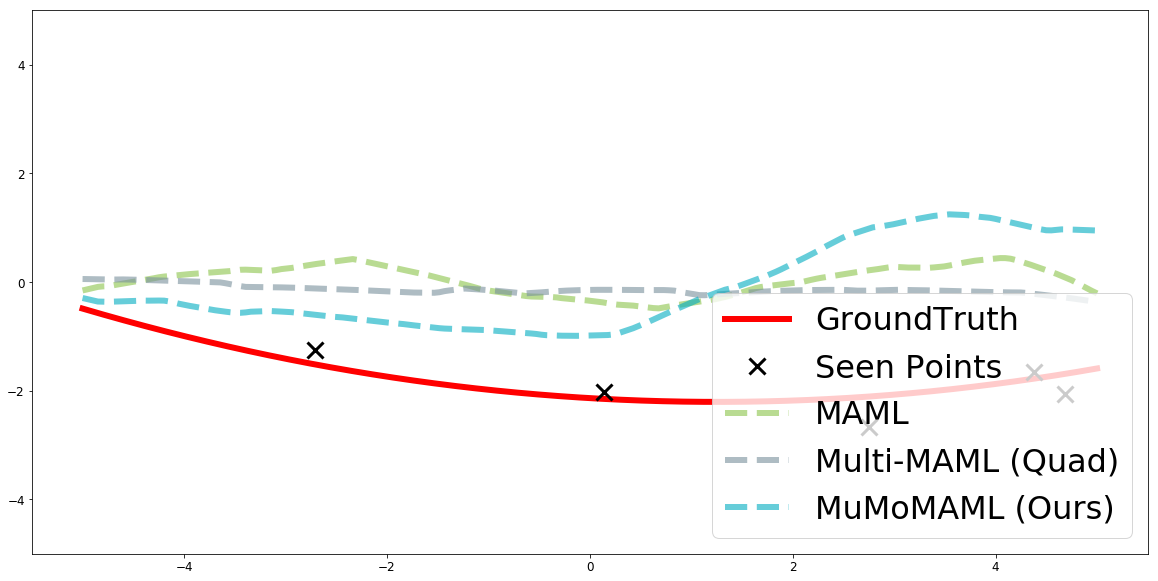} \\
		
		\includegraphics[width=0.305\textwidth,trim={0.8cm 0 0.2cm 0},clip]{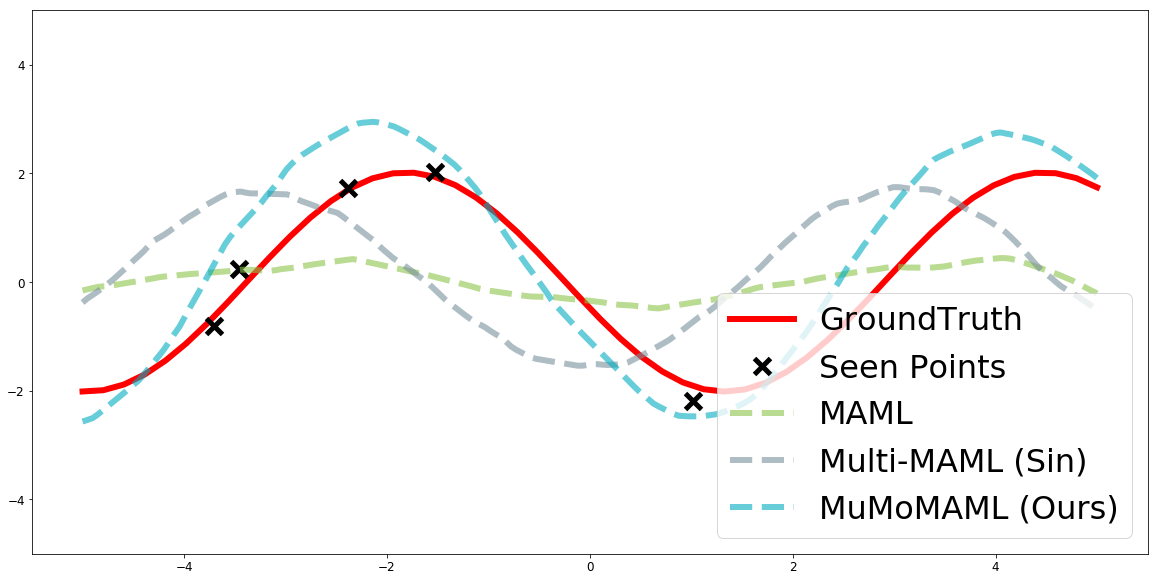} &
		\includegraphics[width=0.305\textwidth,trim={0.8cm 0 0.2cm 0},clip]{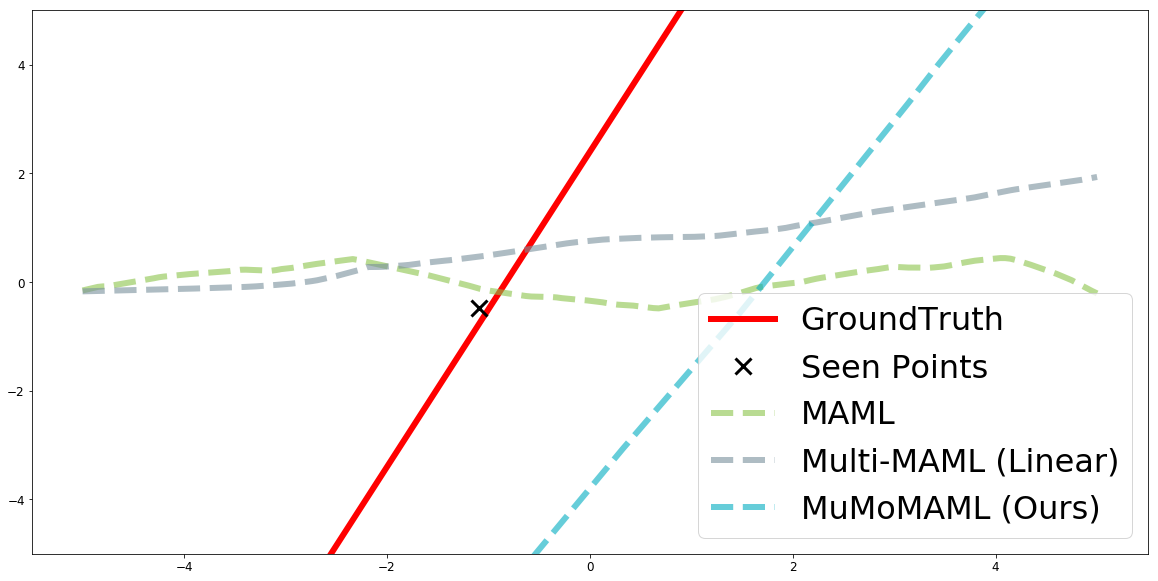} &
		\includegraphics[width=0.305\textwidth,trim={0.8cm 0 0.2cm 0},clip]{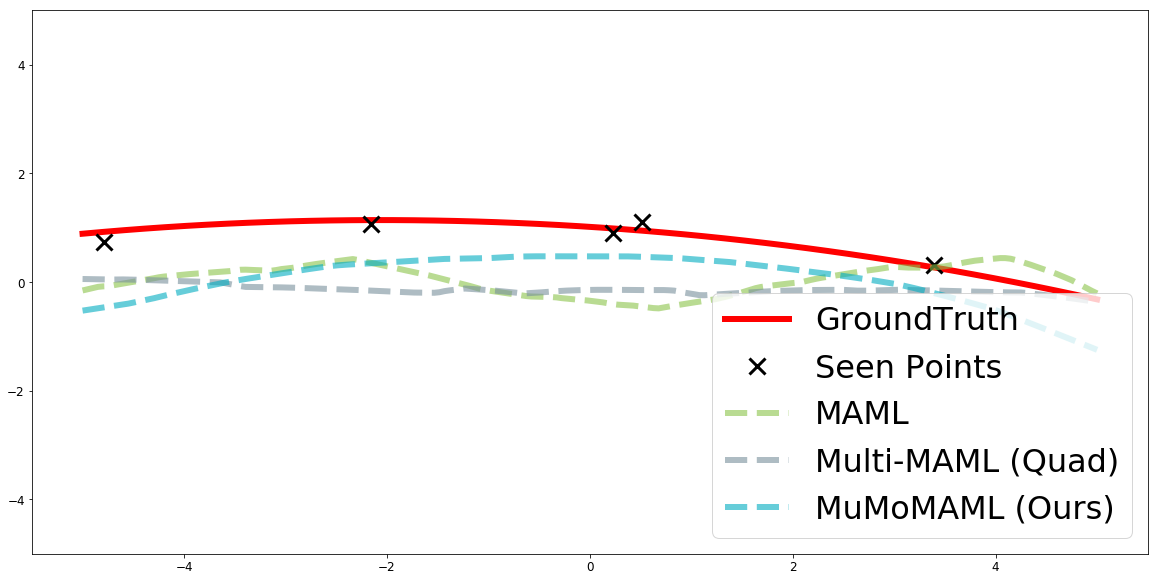} \\
		
		\includegraphics[width=0.305\textwidth,trim={0.8cm 0 0.2cm 0},clip]{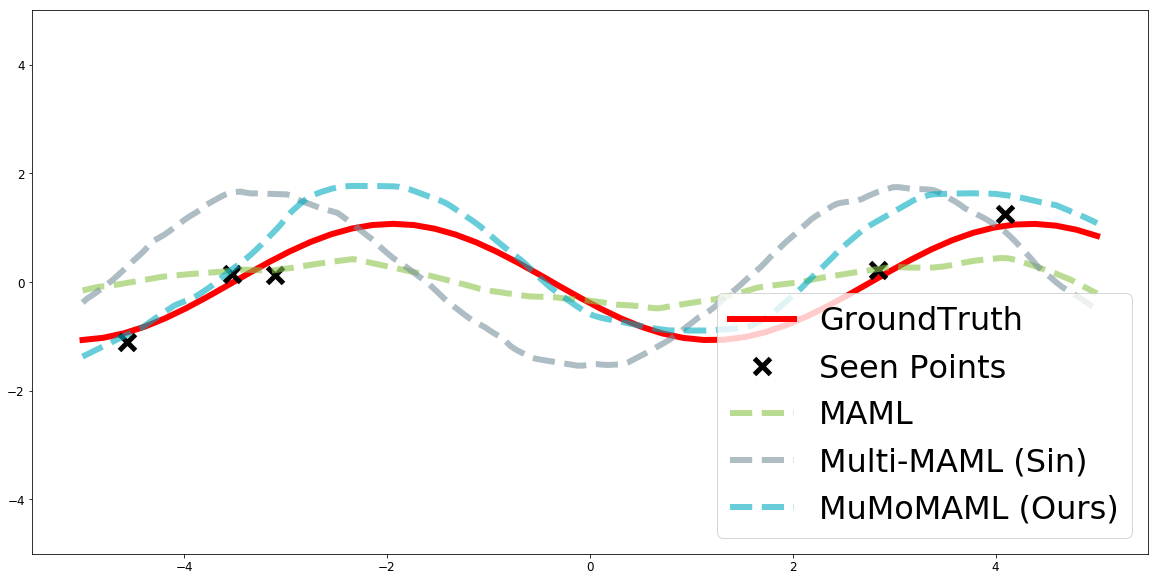} &
		\includegraphics[width=0.305\textwidth,trim={0.8cm 0 0.2cm 0},clip]{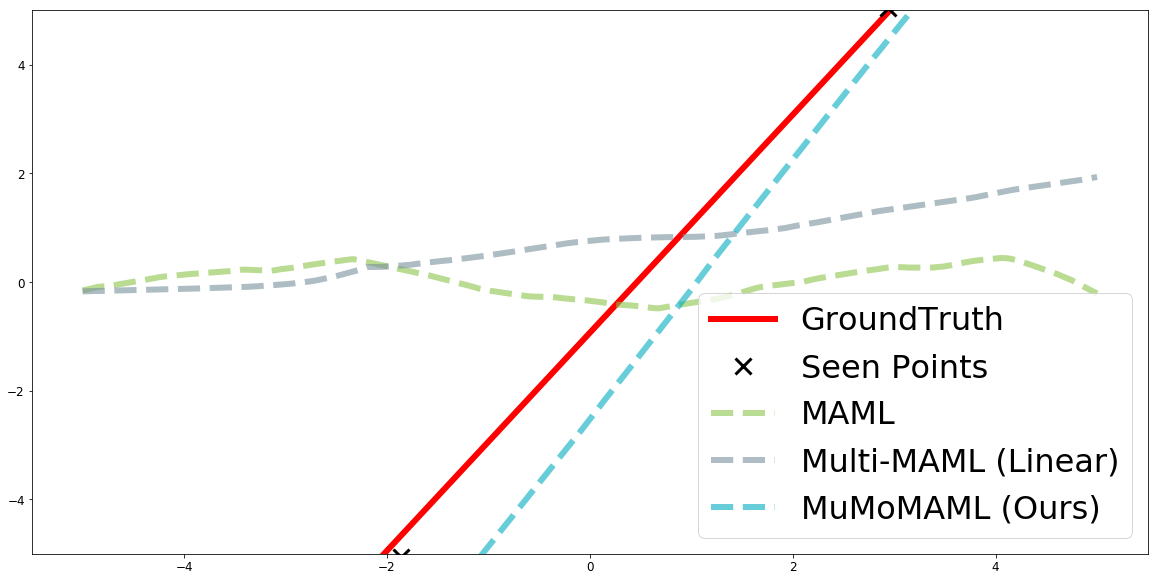} &
		\includegraphics[width=0.305\textwidth,trim={0.8cm 0 0.2cm 0},clip]{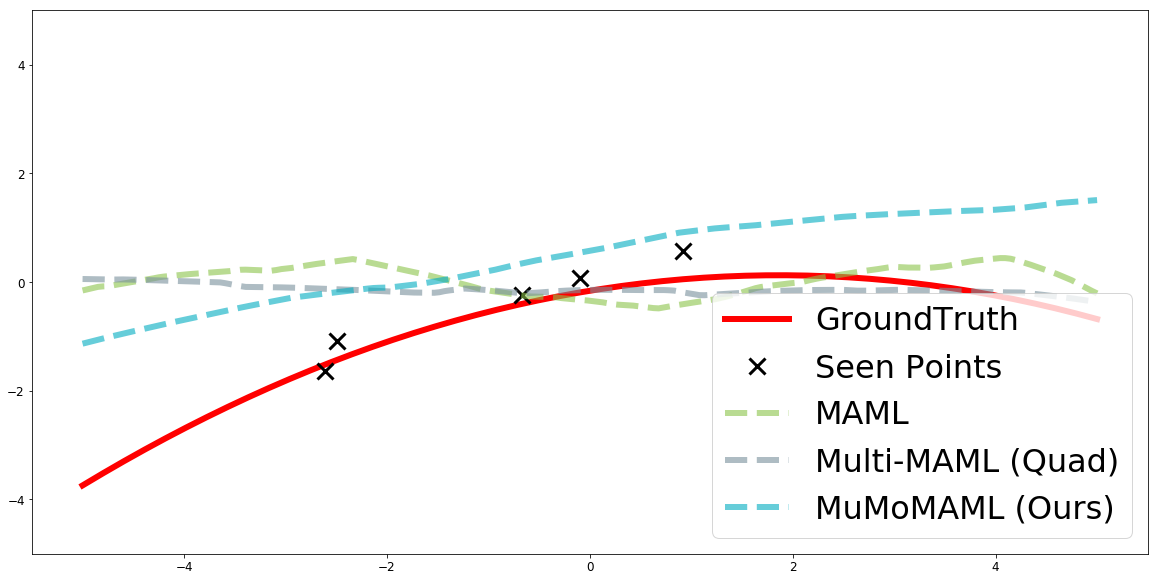} \\
		
		\includegraphics[width=0.305\textwidth,trim={0.8cm 0 0.2cm 0},clip]{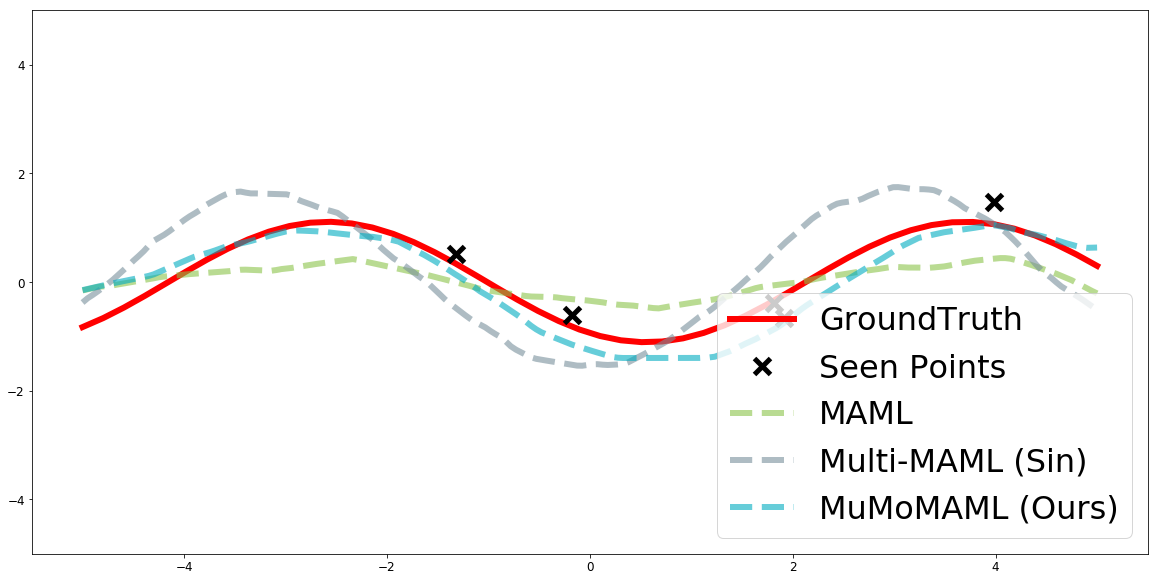} &
		\includegraphics[width=0.305\textwidth,trim={0.8cm 0 0.2cm 0},clip]{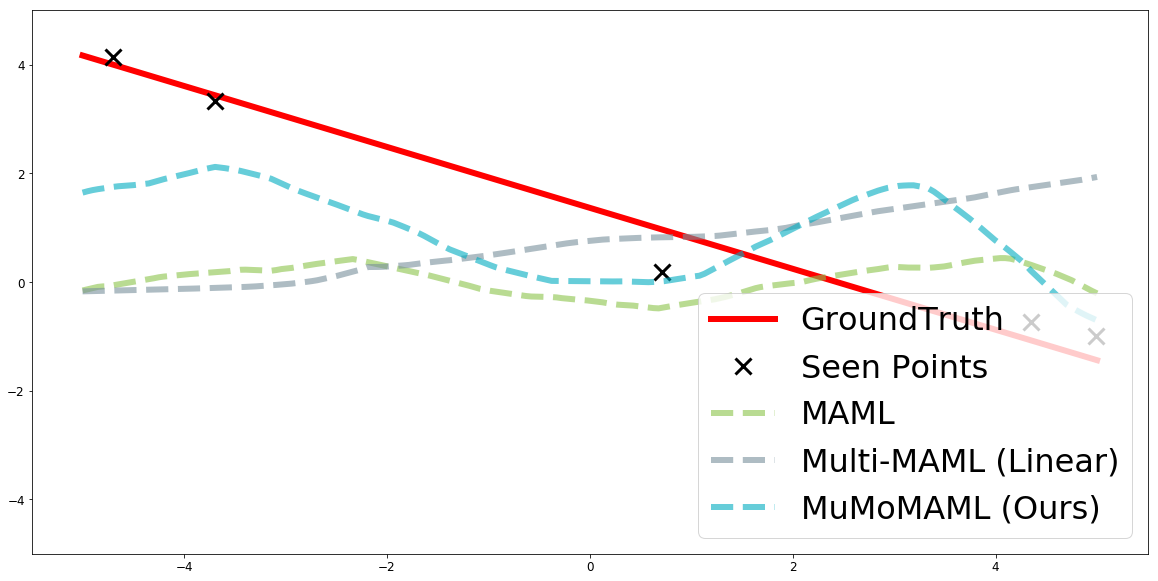} &
		\includegraphics[width=0.305\textwidth,trim={0.8cm 0 0.2cm 0},clip]{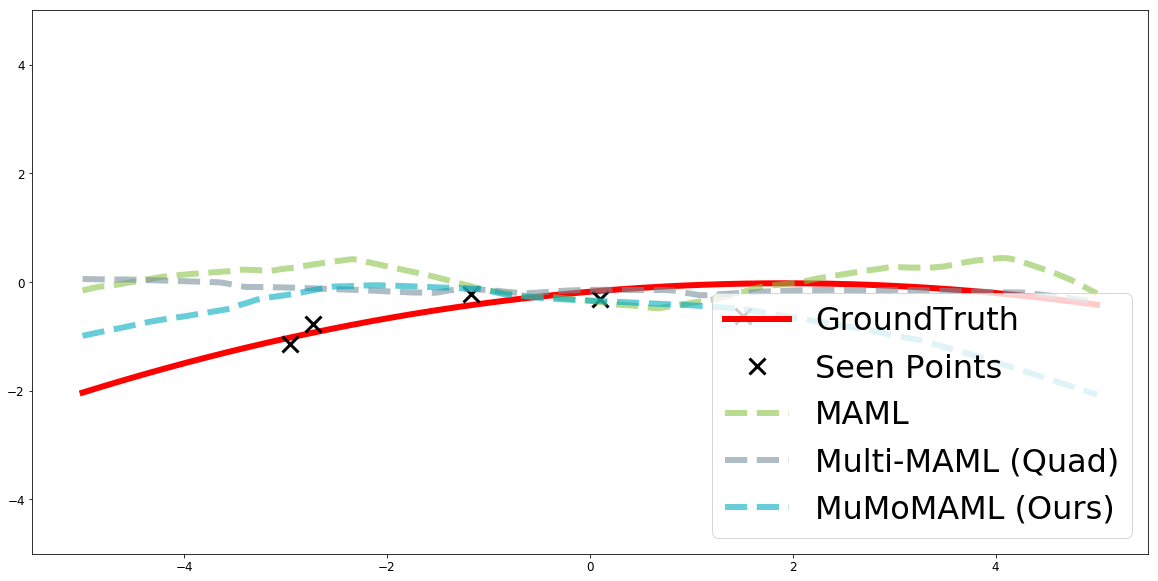} \\
	\end{tabular}
	\caption{
    	\small
	    Additional qualitative results of the regression tasks. \OurMethod \textbf{after modulation} vs. other prior models. 
	}
	\label{fig:regression_A}
\end{figure}

\begin{figure}[ht]
	\centering
	\small
	\begin{tabular}{ccc}
	     Sinusoidal Functions & Linear Functions & Quadratic Functions \\
		\includegraphics[width=0.305\textwidth,trim={0.8cm 0 0.2cm 0},clip]{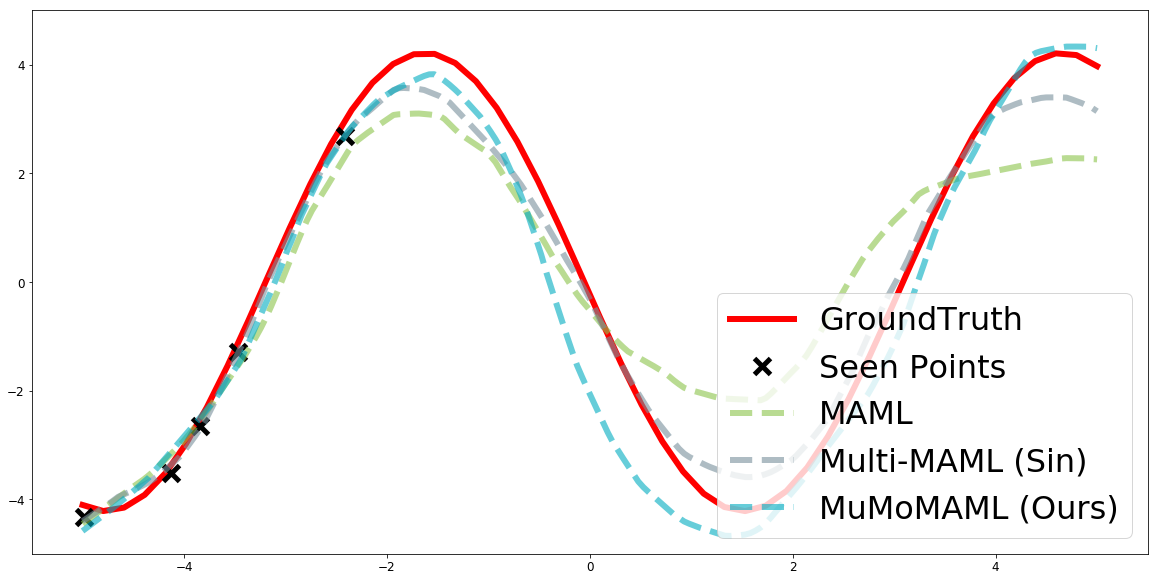} &
		\includegraphics[width=0.305\textwidth,trim={0.8cm 0 0.2cm 0},clip]{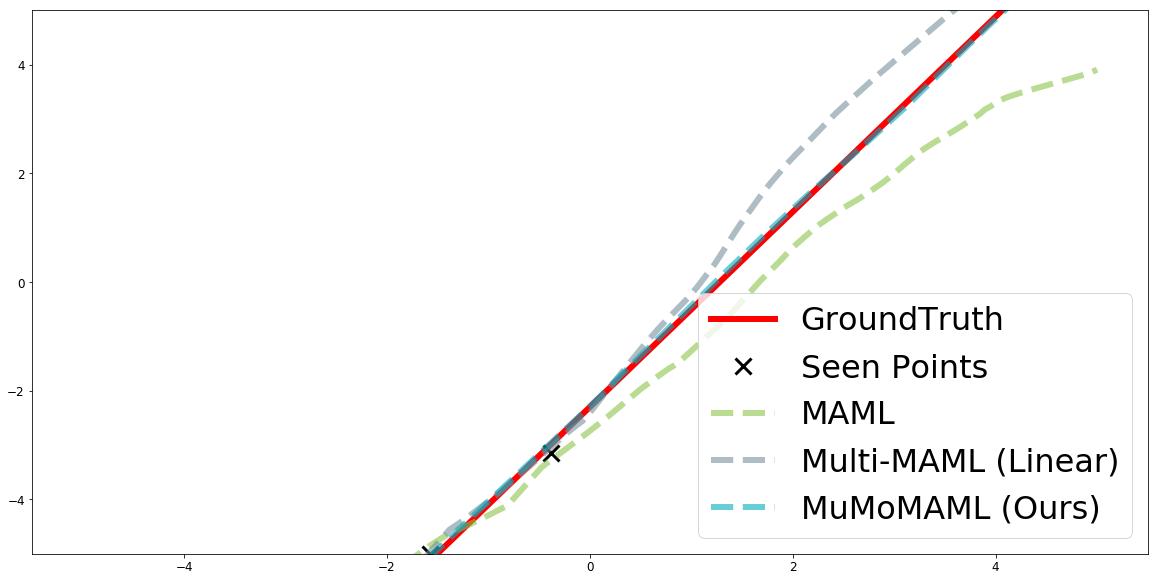} &
		\includegraphics[width=0.305\textwidth,trim={0.8cm 0 0.2cm 0},clip]{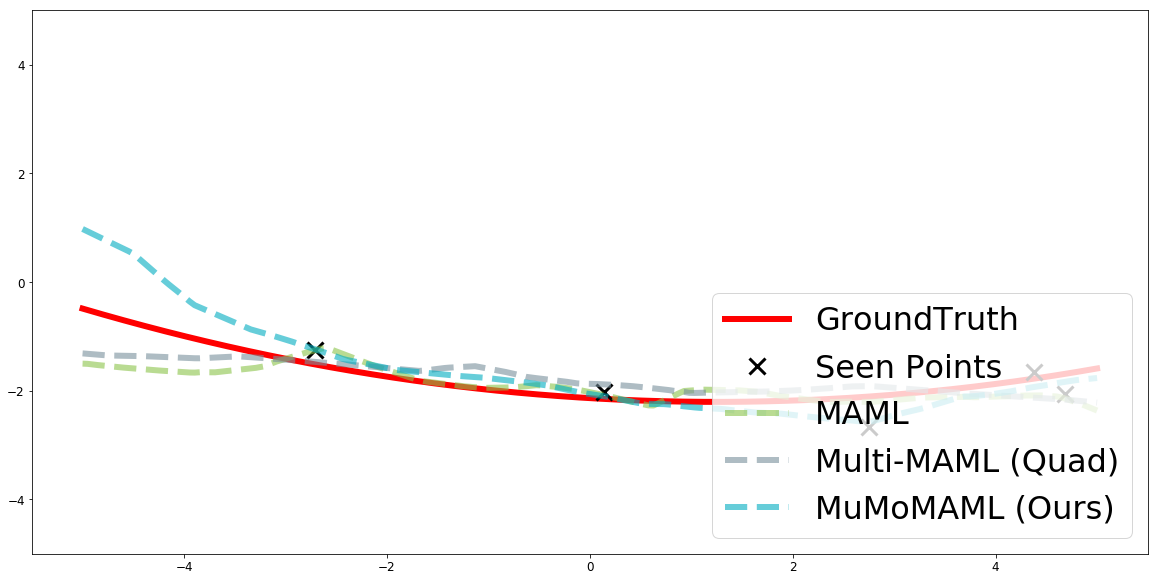} \\
		
		\includegraphics[width=0.305\textwidth,trim={0.8cm 0 0.2cm 0},clip]{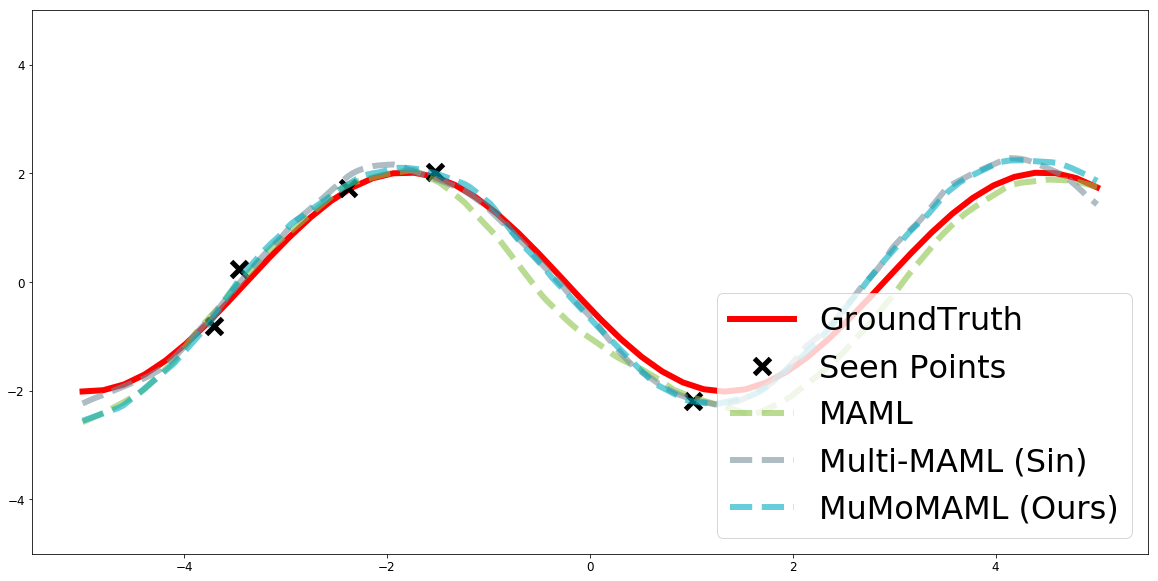} &
		\includegraphics[width=0.305\textwidth,trim={0.8cm 0 0.2cm 0},clip]{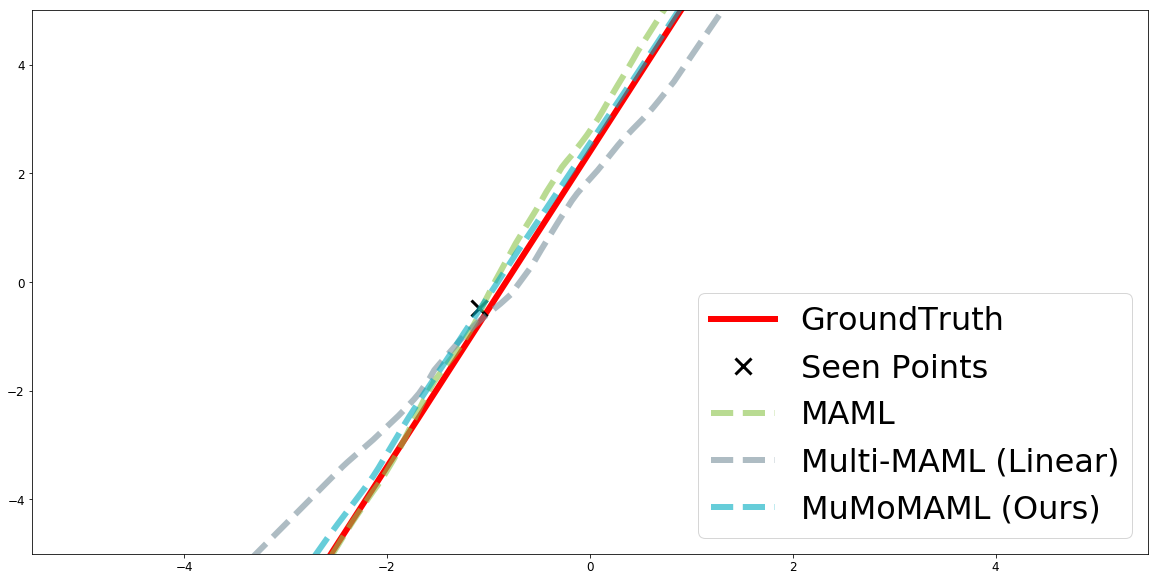} &
		\includegraphics[width=0.305\textwidth,trim={0.8cm 0 0.2cm 0},clip]{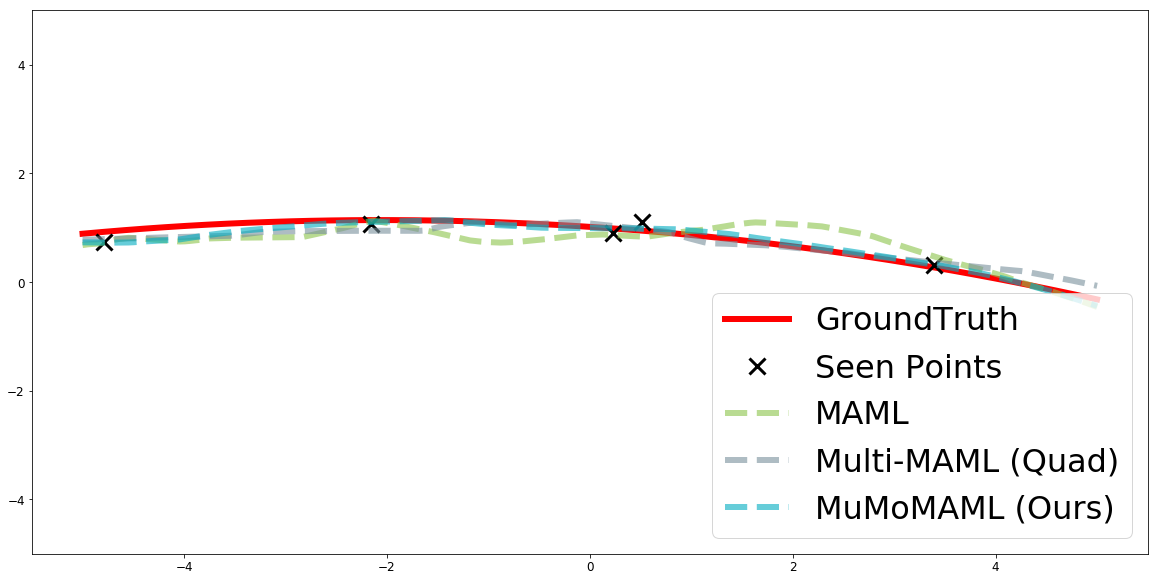} \\
		
		\includegraphics[width=0.305\textwidth,trim={0.8cm 0 0.2cm 0},clip]{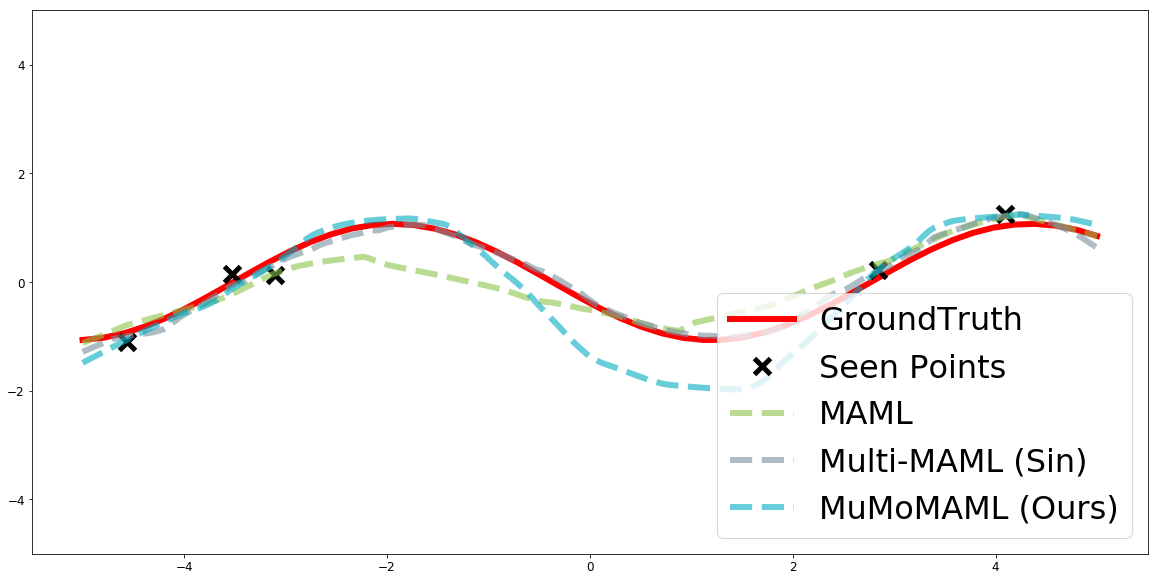} &
		\includegraphics[width=0.305\textwidth,trim={0.8cm 0 0.2cm 0},clip]{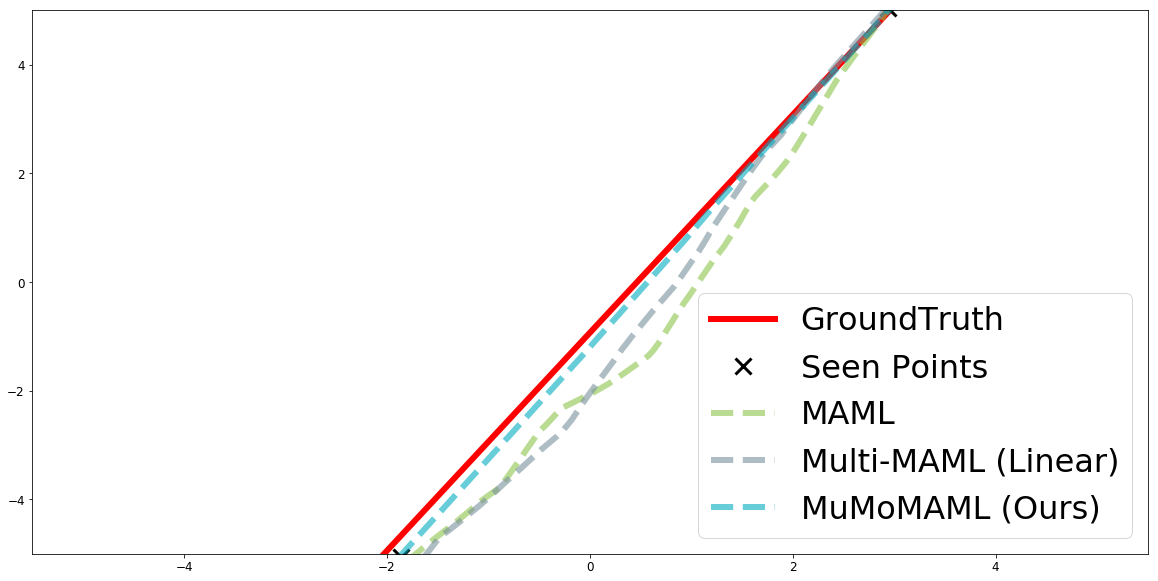} &
		\includegraphics[width=0.305\textwidth,trim={0.8cm 0 0.2cm 0},clip]{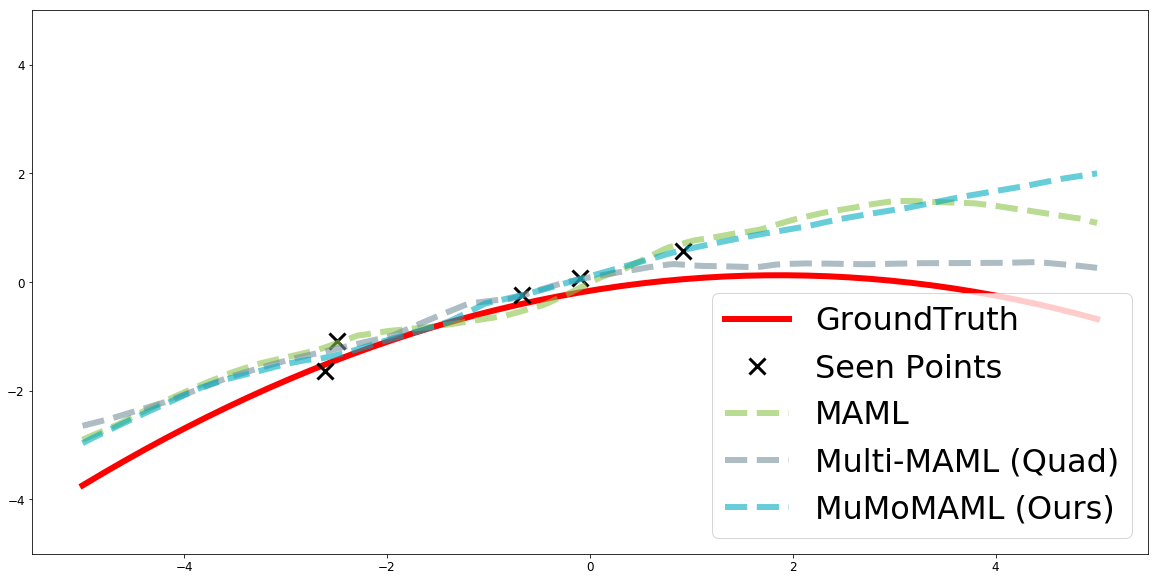} \\
		
		\includegraphics[width=0.305\textwidth,trim={0.8cm 0 0.2cm 0},clip]{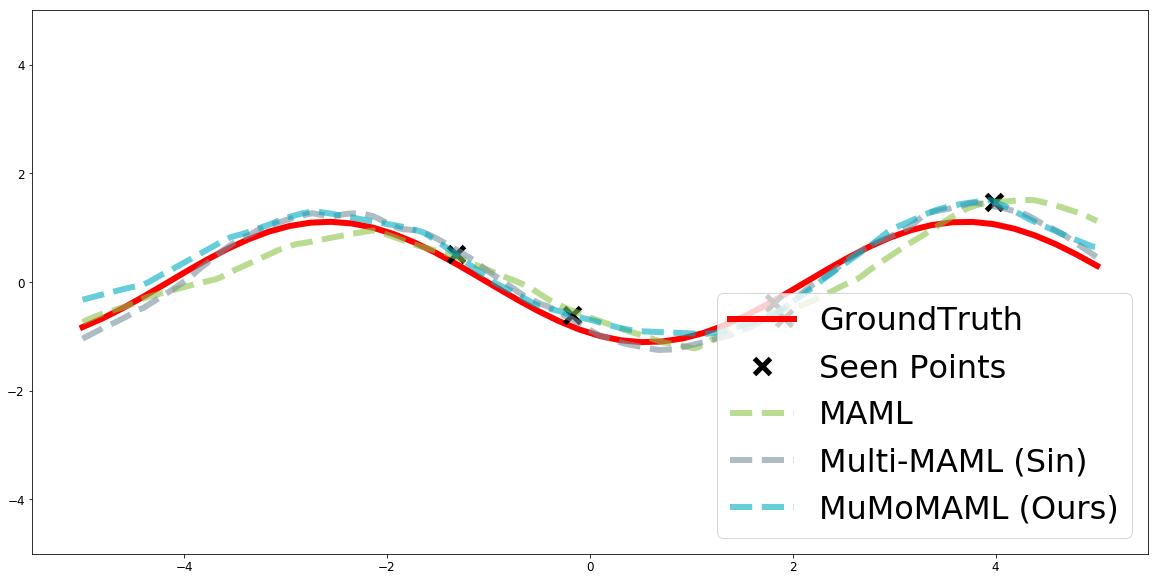} &
		\includegraphics[width=0.305\textwidth,trim={0.8cm 0 0.2cm 0},clip]{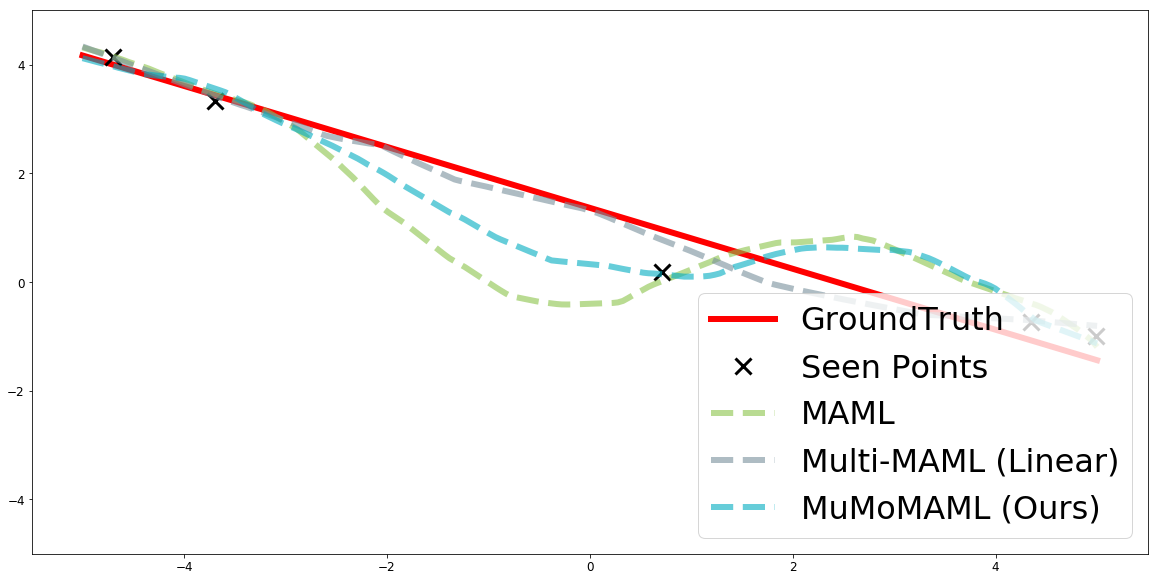} &
		\includegraphics[width=0.305\textwidth,trim={0.8cm 0 0.2cm 0},clip]{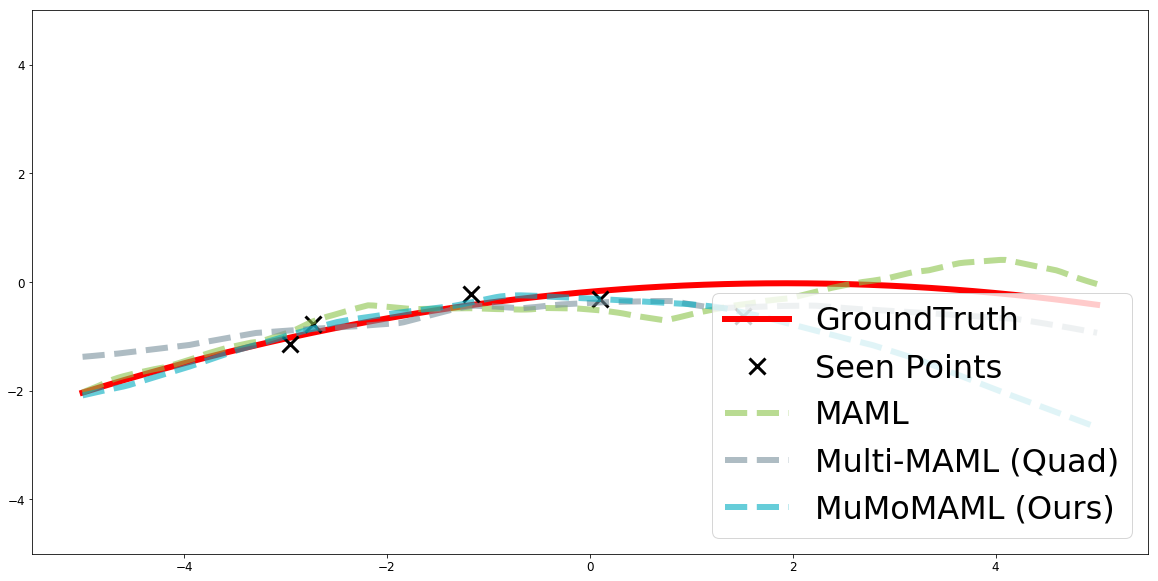} \\
	\end{tabular}
	\caption{
    	\small
	    Additional qualitative results of the regression tasks. \OurMethod \textbf{after adaptation} vs. other posterior models. 
	}
	\label{fig:regression_B}
\end{figure}

Additional qualitative results for \OurMethod after modulation are shown in \myfig{fig:regression_A} and additional qualitative results for \OurMethod after adaptation are shown in \myfig{fig:regression_B}.

\subsection{Additional Qualitative Results for Reinforcement Learning}
\begin{figure}
    \centering
    \includegraphics[width=\textwidth]{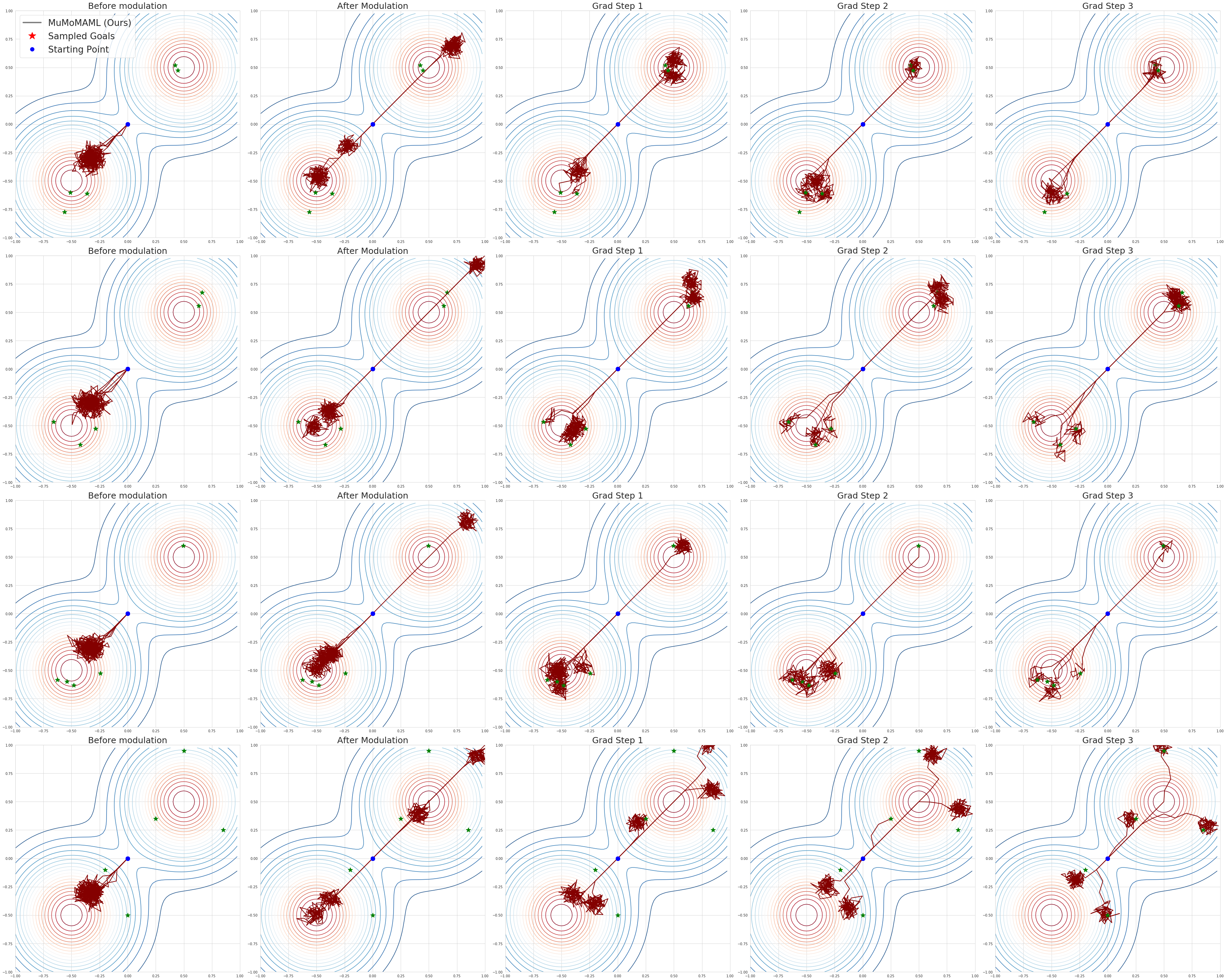}
    \caption{Additional trajectories sampled from the 2D navigation environment with \OurMethod. The first four rows are with goals sampled from the environment distribution, where \OurMethod demonstrates rapid adaptation and is often able to locate the goal exactly. On the fifth row, trajectories are sampled with less probable goals. The agent is left farther away from the goals after the modulation step, but the gradient based adaptation steps then steadily recover the performance.}
    \label{fig:rl_extra_results}
\end{figure}

Additional trajectories sampled from the 2D navigation environment are presented in \myfig{fig:rl_extra_results}.

\end{document}